\documentclass[lettersize,journal]{IEEEtran}
\usepackage{amsmath,amsfonts}
\usepackage{algorithmic}
\usepackage{algorithm}
\usepackage{array}
\usepackage[font=normalsize,labelfont=sf,textfont=sf]{subcaption}
\usepackage{textcomp}
\usepackage{stfloats}
\usepackage{url}
\usepackage{verbatim}
\usepackage{graphicx}
\usepackage{cite}
\usepackage{cleveref}
\usepackage{booktabs}
\usepackage{multicol}
\usepackage{multirow}

\usepackage{xspace}
\usepackage{marvosym}

\makeatletter
\DeclareRobustCommand\onedot{\futurelet\@let@token\@onedot}
\def\@onedot{\ifx\@let@token.\else.\null\fi\xspace}

\def\eg{\emph{e.g}\onedot} 
\def\ie{\emph{i.e}\onedot}

\def\etal{\emph{et al}\onedot}
\makeatother

\crefname{figure}{Fig.}{Figs.}
\Crefname{figure}{Figure}{Figures}
\crefname{table}{Table}{Tables}
\crefname{equation}{Eq.}{Eqs.}
\crefname{section}{Section}{Sections}

\begin{document}

\title{AttenCraft: Attention-based Disentanglement of Multiple Concepts for Text-to-Image Customization}

\author{Junjie Shentu, Matthew Watson, Noura Al Moubayed
\thanks{Junjie Shentu, Matthew Watson, and Noura Al Moubayed are with the Department of Computer Science, Durham University, DH1 3LE Durham, U.K. (Email: noura.al-moubayed@durham.ac.uk\textsuperscript{\Letter}).}}



\maketitle

\begin{abstract}
Text-to-image (T2I) customization empowers users to adapt the T2I diffusion model to new concepts absent in the pre-training dataset. On this basis, capturing multiple new concepts from a single image has emerged as a new task, allowing the model to learn multiple concepts simultaneously or discard unwanted concepts. However, multiple-concept disentanglement remains a key challenge. Existing disentanglement models often exhibit two main issues: feature fusion and asynchronous learning across different concepts. To address these issues, we propose \textbf{AttenCraft}, an attention-based method for multiple-concept disentanglement. Our method uses attention maps to generate accurate masks for each concept in a single initialization step, aiding in concept disentanglement without requiring mask preparation from humans or specialized models. Moreover, we introduce an adaptive algorithm based on attention scores to estimate sampling ratios for different concepts, promoting balanced feature acquisition and synchronized learning. AttenCraft also introduces a feature-retaining training framework that employs various loss functions to enhance feature recognition and prevent fusion. Extensive experiments show that our model effectively mitigates these two issues, achieving state-of-the-art image fidelity and comparable prompt fidelity to baseline models.
\end{abstract}


\section{Introduction}
\label{sec:intro}

\begin{figure*}[htb!]
  \centering
    \includegraphics[width=\linewidth]{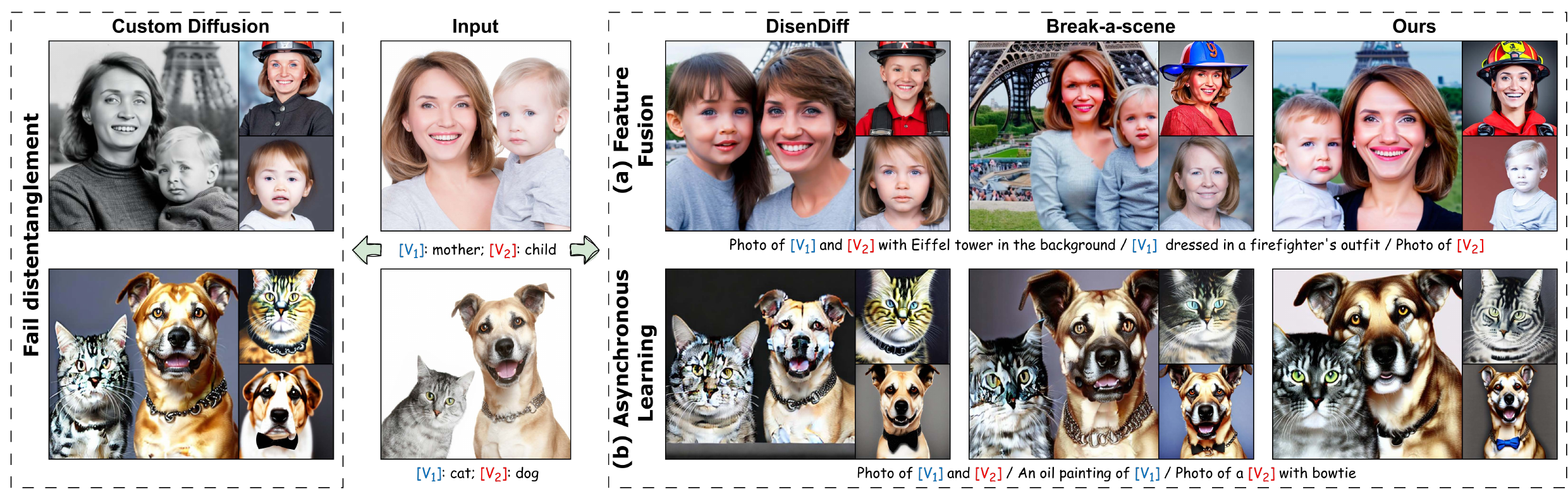}
   \caption{We propose \textit{AttenCraft}, an optimized method for disentangling multiple concepts in a single image. Baseline models present two key issues: (a) feature fusion; (b) asynchronous learning. Our method significantly mitigates these issues and realizes robust concept disentanglement and feature learning.}
   \label{fig:1}
\end{figure*}

\IEEEPARstart{D}{iffusion} models have shown exceptional capabilities in generating high-quality and diverse images \cite{ho2020denoising, dhariwal2021diffusion}. Text-to-image (T2I) diffusion models, in particular, display notable proficiency in producing images aligned with natural language prompts \cite{nichol2021glide, rombach2022high, ramesh2022hierarchical, gu2022vector}. However, incorporating new concepts absent from pre-training datasets remains a challenge \cite{gal2022image}. Studies on ``customizing" T2I models for generalization to new concepts suggested fine-tuning pre-trained models using a few or even a single image of the target object, resulting in subject-driven T2I models \cite{gal2022image, ruiz2023dreambooth, kumari2023multi, li2023blip, jia2023taming, gal2023encoder, arar2023domain, ruiz2023hyperdreambooth, ma2023subject}. In subject-driven T2I learning, the visual representation is mapped to an identifier token $\rm [V]$ via the cross-attention mechanism and is generalized to diverse contexts \cite{gal2022image}. Nonetheless, existing subject-driven T2I models are primarily designed to learn from images containing a single new concept \cite{kumari2023multi}, struggling to learn multiple concepts from one image, as shown in the results of \textit{Custom Diffusion} (\textit{CusDiff}) in \cref{fig:1}.

Several studies have explored learning multiple concepts from a single image or localized regions of the image \cite{avrahami2023break, jin2023image, rahman2024visual, safaee2023clic, zhang2024attention}. Two main strategies for disentangling multiple concepts have been identified. The first strategy uses masks \cite{avrahami2023break, jin2023image, rahman2024visual, safaee2023clic} to guide cross-attention activation during training, represented by \textit{Break-a-scene} (\textit{BAS}) \cite{avrahami2023break}; while the second directly adjusts cross-attention to focus on different concepts in the given image, represented by \textit{DisenDiff} \cite{zhang2024attention}. However, \textit{BAS} depends on masks provided by specialized segmentation models (\eg, SAM \cite{kirillov2023segment}) or human input, while \textit{DisenDiff} struggles to remove background features from the target concepts. More importantly, two key issues that deteriorate the results of concept disentanglement emerge, as presented in \cref{fig:1}. First, baseline models may present feature fusion when learning multiple concepts (\eg, the human haircuts and faces in \cref{fig:1}(a)). Second, an asynchronous learning across different concepts happens in baseline models, as reflected by the ``corruption'' shown in \cref{fig:1}(b). The corruption manifests as noisy patches, which indicates overfitting \cite{wu2024exploring} of the corresponding concept. The asynchronous learning can be observed between the single concept and concept group (\textit{DisenDiff}), and between different single concepts (\textit{BAS}), depending on specific model settings. A detailed analysis will be presented in \cref{sec3.3}.

In this study, we propose \textit{AttenCraft}, a novel method for disentangling multiple concepts from a single image in subject-driven T2I generation. Specifically, we adopt the mask-based strategy for disentanglement, using self-attention and cross-attention maps to generate accurate masks for each target concept in a single step, without the need for specialized segmentation models or human input. These masks guide cross-attention activation for disentanglement during training. Aligning the cross-attention map of the identifier token $\rm [V]$ with the corresponding mask establishes an explicit connection between $\rm [V]$ and the visual representation of the target concept. We also investigate the relationship between feature acquisition and the initialization of $\rm [V]$, proposing an adaptive algorithm that automatically estimates the sampling ratio of multiple concepts based on cross-attention scores. This approach mitigates asynchronous learning and enhances learning quality. Furthermore, we demonstrate that back-propagating reconstruction loss during multiple-concept sampling is a primary cause of feature fusion. Thus, we optimize the training framework by introducing different loss functions for sampled subsets with varying sizes. Overall, our contributions are threefold:

\begin{itemize}
    \item We leverage the cross-attention and self-attention maps to create precise masks for each concept in a given image within a single initialization step, without relying on specialized models or human input;
    \item We propose an adaptive algorithm that automatically estimates the sampling ratio of multiple concepts based on cross-attention scores, mitigating the issue of asynchronous learning;
    \item We introduce a novel feature-retaining training framework that applies different loss functions to sampled subsets of varying sizes, effectively preventing feature fusion and improving the quality of feature acquisition.
\end{itemize}

\section{Related Work}
\label{sec:related_work}
\subsection{Diffusion models and T2I customization} By utilizing pre-trained text encoders \cite{vaswani2017attention, ramesh2021zero}, diffusion models implement the T2I diffusion model in pixel space under classifier-free guidance \cite{nichol2021glide, saharia2022photorealistic, ho2022classifier}. Stable Diffusion (SD) \cite{rombach2022high} trains the denoising UNet \cite{ronneberger2015u} in latent space by applying a Variational Autoencoder (VAE) \cite{kingma2013auto} and the text encoder of Contrastive Language-Image Pre-training (CLIP) \cite{radford2021learning} model. Furthermore, subject-driven T2I models \cite{gal2022image, ruiz2023dreambooth} learn a new concept from several images and reverse to an identifier token $\rm [V]$. In addition, parameter-efficient tuning (PEFT) \cite{cao2024controllable} is employed to minimize training time by utilizing a smaller set of trainable parameters. These include cross-attention layers \cite{kumari2023multi, zhang2023ssr, zhang2024attention, cai2024decoupled}, Low-rank Adaptation (LoRA \cite{hu2021lora}) parameters \cite{ruiz2023hyperdreambooth, gu2024mix, chen2023disenbooth, yang2024lora}, and supplementary components such as an encoder, adapter, or weight offset \cite{ hao2023vico, gal2023encoder, liu2023cones}. Moreover, some studies pre-train a universal encoder capable of directly encoding input images \cite{ma2023subject, li2023blip,shi2023instantbooth, wei2023elite, chen2024subject}. However, the majority of subject-driven T2I models focus on input images containing a single concept, neglecting the exploration of extracting multiple concepts from a single image.

\subsection{Application of attention in diffusion models} The Attention mechanism manipulates feature dependencies during T2I generation. Guided by cross-attention, pre-trained diffusion models exhibit superior semantic alignment with provided text prompts \cite{feng2022training, chefer2023attend, wang2023compositional, phung2023grounded}, achieve image editing \cite{hertz2022prompt, nguyen2024flexedit}, and provide positional control \cite{ma2023directed, he2023localized, chen2023training, phung2023grounded, liu2023cones}. Moreover, cross-attention guidance is applied during model training to eliminate background interference or concentrate on specific regions in input images using provided masks \cite{ma2023subject, wei2023elite, hao2023vico, chen2023disenbooth, avrahami2023break, safaee2023clic, shentu2024textual}. Meanwhile, self-attention can promote subject consistency across different contexts \cite{tewel2024training} or facilitate subject swaps while preserving style consistency \cite{jeong2024visual}. Furthermore, the self-attention and cross-attention maps are applied to achieve unsupervised segmentation \cite{tian2023diffuse} and augmentation of the segmentation datasets \cite{wu2023diffumask, nguyen2024dataset, marcos2024open}.

\subsection{Disentangling multiple concepts from a single image} \textit{BAS} disentangles multiple concepts from a single image by applying masks provided by users or specialized segmentation models \cite{kirillov2023segment} to guide cross-attention activation. Meanwhile, Saffee \etal \cite{safaee2023clic} adopt automatically identified masks to learn a given concept and apply them to edit other images. Jin \etal \cite{jin2023image} apply a fixed threshold on the cross-attention maps to obtain the mask. Furthermore, Rahman \etal \cite{rahman2024visual} utilize dense conditional random field (CRF) \cite{krahenbuhl2011efficient}, and Hao \etal \cite{hao2023vico} applies Otsu thresholding \cite{otsu1975threshold}, to obtain masks from cross-attention maps. However, these automatic masks are typically coarse \cite{jin2023image, rahman2024visual}, time-consuming \cite{safaee2023clic}, and often fail to separate different concepts \cite{hao2023vico}. \textit{DisenDiff} \cite{zhang2024attention} calibrates cross-attention to encourage the model to separate its attention and achieve disentanglement without masks, but fails to exclude the background. Our proposed approach efficiently disentangles multiple concepts and backgrounds from a single input image using self-generated accurate masks guided by the attention mechanism.

\section{Proposed Method}
\label{method}

\begin{figure*}[htb!]
  \centering
    \includegraphics[width=1.0\linewidth]{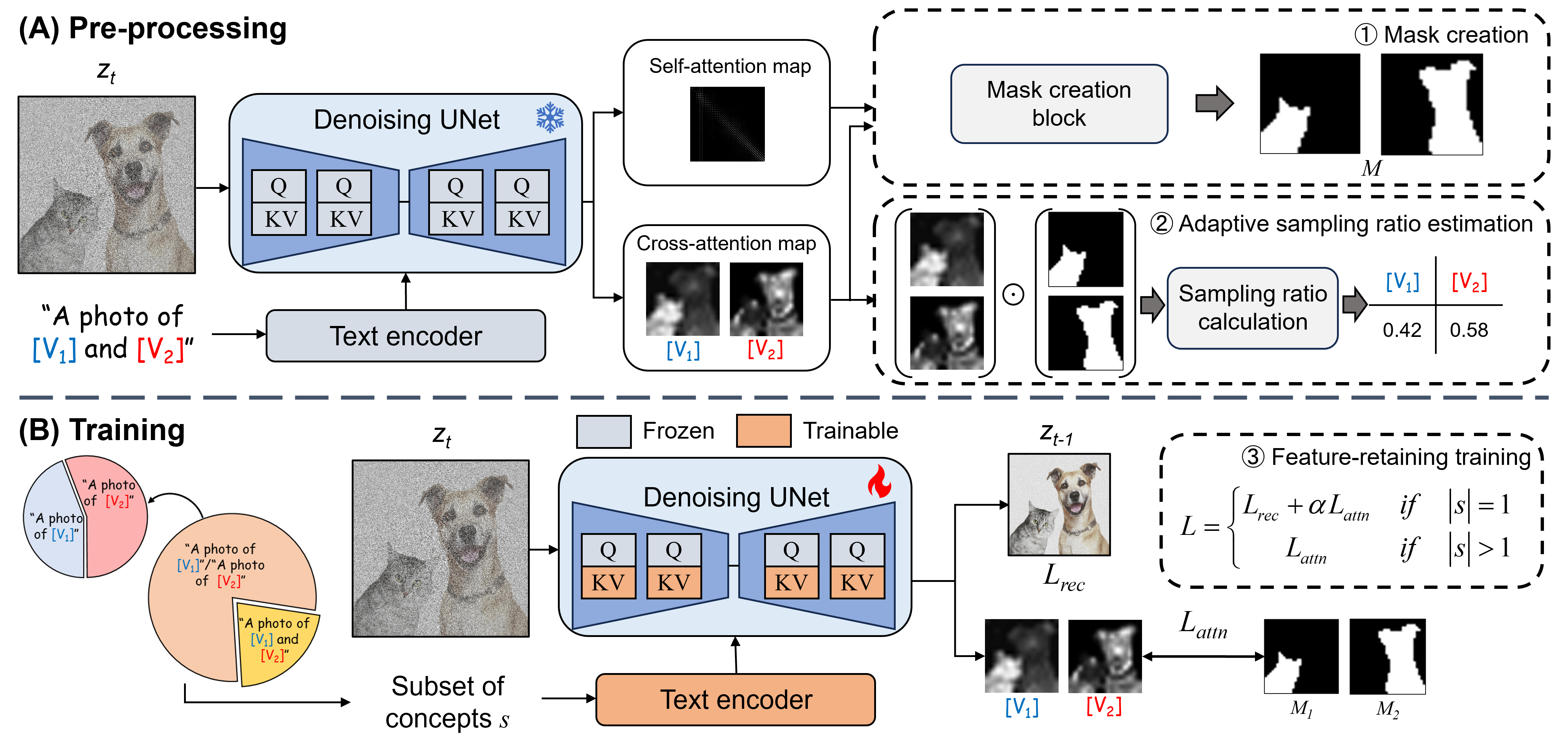}

   \caption{\textbf{Method overview.} Given an image with multiple concepts, within a few steps in the pre-processing stage, we create accurate masks for each concept and adaptively estimate the sampling ratio for multiple concepts to enhance learning synchronicity. We also propose an optimized training framework by introducing different loss functions for sampled subsets of varying sizes to prevent feature fusion.}
   \label{fig:2}
\end{figure*}

In this section, we begin by providing a brief overview of the diffusion model, and then introduce our method, which includes mask auto-creation guided by attention maps, adaptive estimation of sampling ratios of different concepts, and a dedicated training framework to prevent feature fusion across concepts. An illustration of our method is presented in \cref{fig:2}.

\subsection{Preliminary}
For an input image $x \in \mathbb{R}^{H\times W\times 3}$, SD projects $x$ into a latent representation $z \in \mathbb{R}^{h\times w\times c}$ via a VAE encoder $\mathcal{E}$ \cite{kingma2013auto}, where $c$ is the latent feature dimension. The text prompts $y$ are projected into text embeddings by a pre-trained CLIP text encoder $\tau_\theta$. The UNet is trained to predict the randomly added noise $\varepsilon$ given the noisy latent $z_t$, the timestep $t$, and conditioning: 
\begin{equation}
\label{eq:1}
    L_{LDM} = \mathbb{E}_{z,y,t, \varepsilon} \left [ \left \| \varepsilon - \varepsilon_{\theta} \left ( z_t,t,\tau_\theta(y) \right )  \right \|^2_2 \right ] 
\end{equation}

\noindent where $\varepsilon$ and $\varepsilon_\theta$ are standard Gaussian noise and predicted noise residual, respectively. The UNet incorporates self-attention and cross-attention layers to capture the dependencies within the input data \cite{vaswani2017attention, rombach2022high}. The self-attention layers capture the global attention within the image while the cross-attention layers learn to attend between the image and text prompts. The cross-attention map $A_C$ and self-attention map $A_S$ can be calculated as follows:

\begin{subequations}\label{eq:2}
\begin{align}
A_C = softmax \left ( Q_I K_T^\top / \sqrt{d} \right ) \label{eq:2A}\\
A_S = softmax \left (Q_I K_I^\top / \sqrt{d} \right ) \label{eq:2B}
\end{align}
\end{subequations}

\noindent where $Q_I$, $K_I$, $K_T$ are the query matrix, key matrix of $z_t$, and key matrix of $\tau_\theta(y)$, respectively.

\subsection{Attention-guided mask creation}
\label{sec3.2}
\begin{figure}[htb]
  \centering
   \includegraphics[width=1.0\linewidth]{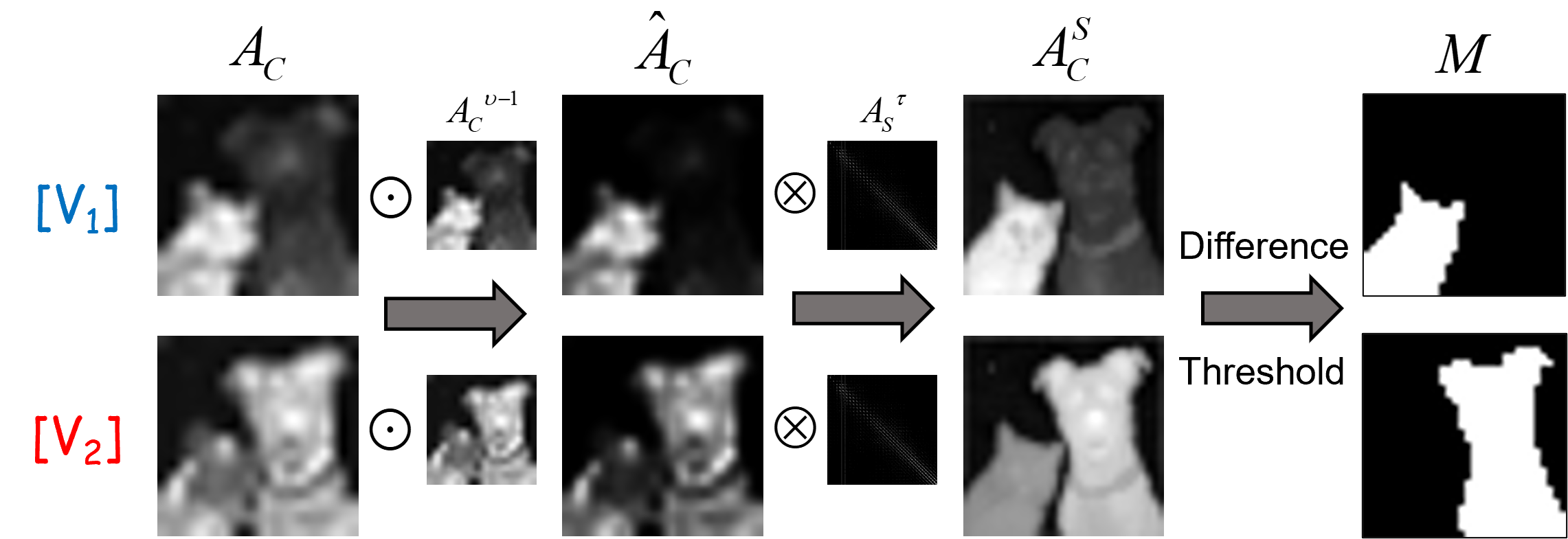}

   \caption{\textbf{Process of attention-guided mask creation.} By applying the cross-attention and self-attention maps, precise masks can be created without specialized models or human inputs.}
   \label{fig:4}
\end{figure}

The cross-attention map $A_C$ outlines the location and shape of the target concept. However, it often displays coarse granularity and noise, leading to two main challenges in mask creation: (1) strong attention activation is shown within the target region, but weak activation occurs in other areas; (2) attention activation is unevenly distributed, leading to an incomplete representation, as shown in \cref{fig:4}. To address the first challenge, we apply \textit{Cross-attention suppression} \cite{zhang2024attention} following the left part of \cref{eq:4}:

\begin{equation}
\label{eq:4}
    \hat{A}_{C} = (A_C)^{\upsilon},A_C^S = \hat{A}_{C} \otimes (A_S) ^ {\tau}
\end{equation}

The activation values of the attention map, generated through a \textit{Softmax} operation, range from 0 to 1. Consequently, element-wise exponentiation of $A_C$ by $\upsilon$ can reduce weak activation in non-target regions but amplifies uneven activation. To address this, we use \textit{Self-attention enhancement} \cite{nguyen2024dataset}, which multiplies $\hat{A}_{C}$ by $A_S^{\tau}$ to enhance the smoothness and precision of $\hat{A}_{C}$, as depicted in the right part of \cref{eq:4}. $A_S$ captures pairwise correlations among patches in $z_t$, allowing attention activation to spread to related regions while reducing activation elsewhere. Similarly, element-wise exponentiation of $A_S$ by $\tau$ decreases correlations between patches of different concepts. With $A_C^S$, we observe that the attention activation of different tokens emphasizes distinct regions in the attention map. Thus, masks can be inferred from activation differences. For the target concept $i$, we compute the maximum difference between its processed attention map $A_{C_i}^S$ and that of another concept $j$ ($A_{C_j}^S$, $i \ne j$), setting the mask value $M_i$ to 1 if it exceeds a preset threshold $\gamma$. We term this process \textit{Delta masking}, and it is defined by the following:
\begin{equation}
\label{4}
    M_i  = \left\{
    \begin{matrix}
    True & \textit{if} \max(A_{C_i}^S   - A_{C_j}^S ) > \gamma, i \neq j \\
    False & \textit{Otherwise}
    \end{matrix}
    \right.
\end{equation}

Attention-guided mask creation is performed within the mask creation block shown in \cref{fig:2}. This process requires only a single step, where the noisy latent $z_t$ is sampled from $t \in [0, 300]$ in the DDPM noise schedule \cite{ho2020denoising} since $z_t$ retains finer semantic details at this stage \cite{nguyen2024flexedit}. Details of the mask creation process are depicted in \cref{fig:4}.

\subsection{Adaptive sampling ratio estimation based on attention scores}
\label{sec3.3}
The issue of asynchronous learning is illustrated in \cref{fig:1}. Wu \etal \cite{wu2024exploring} note that the corruption is caused by a narrowed learning distribution when applying few-shot or one-shot learning, creating a limited window between underfitting and overfitting. The adverse effects of this limited window are pronounced when learning multiple concepts, as the learning windows for different concepts may not align perfectly, leading to asynchronous learning. Baseline models use a fixed sampling scheme during training, which cannot adapt to varied inputs. Specifically, \textit{DisenDiff} utilizes a consistent text prompt encompassing all target concepts throughout the training process, resulting in the overfitting of the concept group when single concepts are properly learned; while \textit{BAS} employs a union sampling scheme that randomly selects a subset of multiple target concepts to form the text prompt, achieving comparatively better learning synchronicity than \textit{DisenDiff}. However, the sampling ratio for each single concept in \textit{BAS} remains identical, which still raises the asynchronous learning issue since the learning steps required for different concepts vary. Thus, an optimized sampling scheme with an adaptive sampling ratio for different concepts is required.

\begin{figure}
  \centering
  \begin{subfigure}{0.46\linewidth} 
    \includegraphics[width=1.0\linewidth]{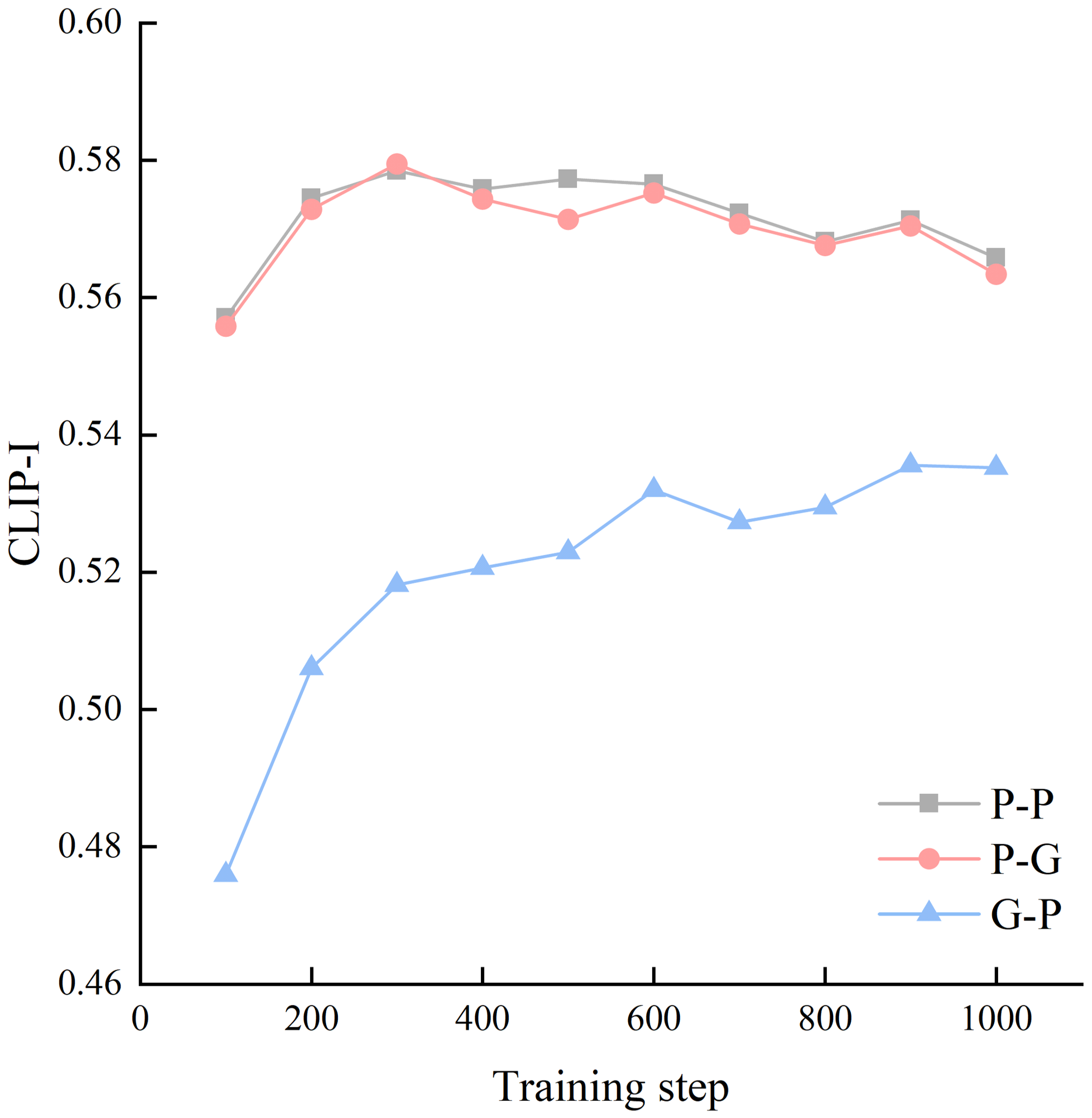}
    \caption{}
    \label{fig:3a}
  \end{subfigure}
  \hfill
  \begin{subfigure}{0.46\linewidth} 
    \includegraphics[width=1.0\linewidth]{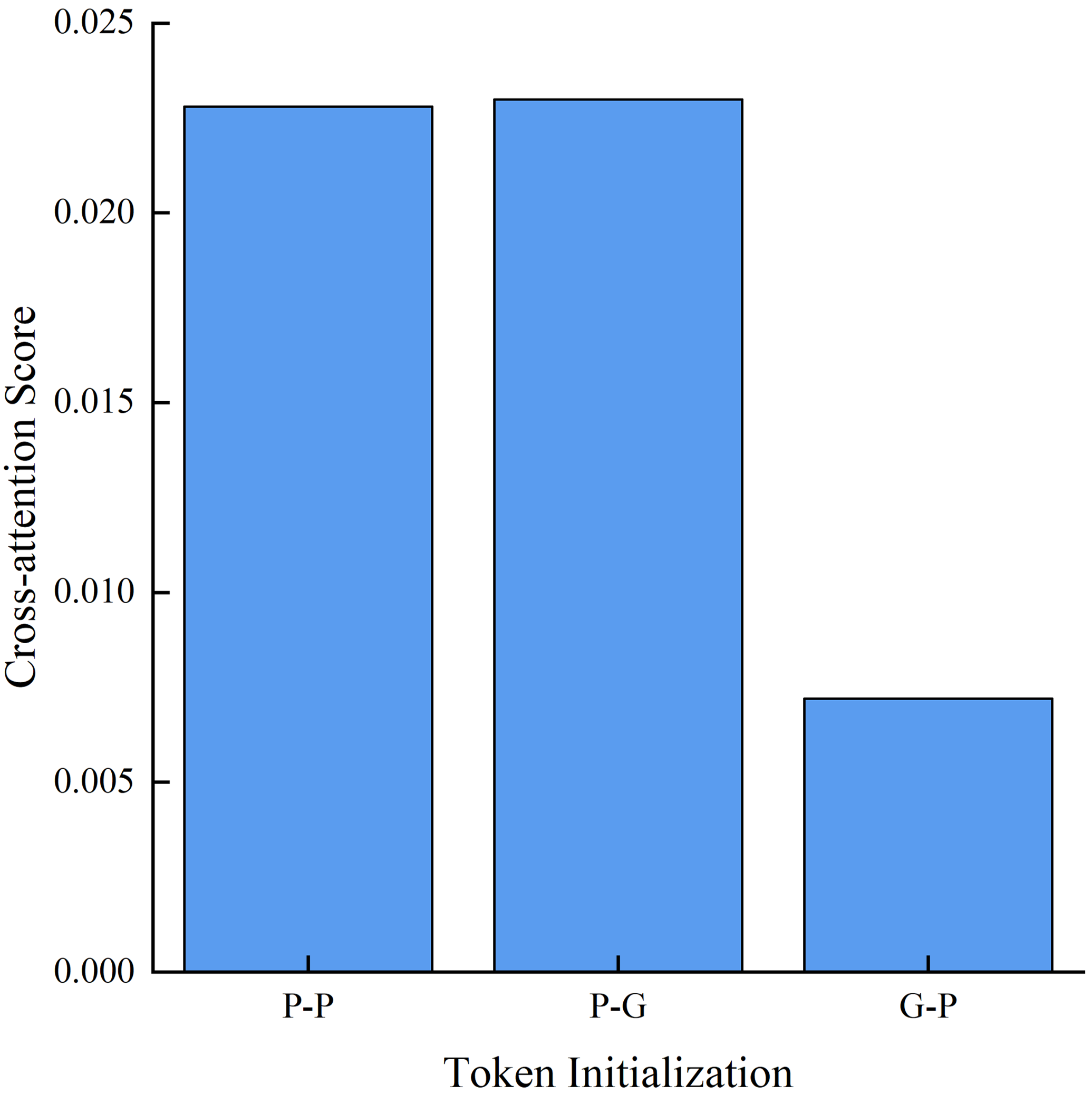}
    \caption{}
    \label{fig:3b}
  \end{subfigure}
  \caption{\textbf{Results of the token initialization experiment.} (a) Variation of single-concept CLIP-I scores with training step; (b) The highest cross-attention score of $\rm [V]$ concerning different initialization patterns.}
  \label{fig:short}
\end{figure}

\paragraph{Identifier token initialization}
We first investigate the relationship between feature acquisition and identifier token initialization through a preliminary experiment, where \textit{BAS} is deployed to learn multiple concepts from 10 datasets \cite{zhang2024attention}, each containing two concepts, over 1000 training steps. Before training, identifier tokens $\rm [V_1]$ and $\rm [V_2]$ are initialized by text embeddings of existing tokens. For each dataset, we apply three token initialization patterns using text embeddings of the precise class (dubbed as \textit{P}) and the general category (dubbed as \textit{G}), resulting in \textit{P}-\textit{P}, \textit{P}-\textit{G}, and \textit{G}-\textit{P}. The CLIP-I scores of the generated images are assessed to reflect the feature acquisition of each concept (detailed in \cref{sec 4.1}). Note that this experiment assesses single-concept generation, meaning that the initialization pattern varies relative to concepts within the same dataset \footnote{For example, in the ``cat \& dog" dataset, we set the triplet ``cat-dog", ``cat-animal", and ``animal-dog" as different initialization patterns. These correspond to \textit{P}-\textit{P}, \textit{P}-\textit{G}, and \textit{G}-\textit{P} when evaluating the ``cat", while \textit{P}-\textit{P}, \textit{G}-\textit{P}, and \textit{P}-\textit{G} when evaluating the ``dog".}. The variation in CLIP-I scores over training steps is shown in \cref{fig:3a}. The model begins at a higher initial point when $\rm [V]$ is initialized with a precise class \textit{P} but tends to degrade after 300 steps. Conversely, when initialized with the general category \textit{G}, the model starts lower but continues to learn until the end of training. These results indicate that the initialization of $\rm [V]$ significantly impacts the feature acquisition.

\paragraph{Attention activation and sampling ratio}
The difference between \textit{P} and \textit{G} for learning lies in their semantic connection to the target concepts, which can be reflected in the cross-attention scores. We validate this hypothesis by extracting the highest activation score from the cross-attention map of each $\rm [V]$, with the result presented in \cref{fig:3b}. The results show that cross-attention scores are significantly higher when a specific concept (\eg, $\rm [V_1]$) is initialized with \textit{P}, while the initialization of the other concept $\rm [V_2]$ has negligible effects on $\rm [V_1]$'s activation score. This observation is further supported by the results in \cref{fig:3a}, where the CLIP-I scores for \textit{P}-\textit{G} and \textit{P}-\textit{P} show only marginal divergence. In conclusion, an identifier token $\rm [V]$ initialized with a less semantically rich embedding requires more steps for feature learning and should be assigned a larger sample ratio to achieve more balanced and synchronized feature acquisition, where the implicit semantic connection can be explicitly reflected by cross-attention scores.

\paragraph{Adaptive sampling ratio estimation}
We propose an attention-based algorithm for an adaptive sampling ratio estimation, grounded in experimental results. Specifically, we first apply self-created masks (see \cref{sec3.2}) on cross-attention maps following $A_M = A_C \odot M$ to eliminate the noise outside the target region, and then extract the highest activation score $S$ from the masked maps. To mitigate contingency, we average the $n$ highest activation scores across $m$ denoising timesteps, as expressed in:

\begin{equation}
\label{eq:7}
    S  = \frac{1}{m} \sum_{t\in \mathbb{T}} \frac{1}{n}\sum_{i=1}^{n}\max A_{M_t}^{(k)}
\end{equation}

\noindent where $\max A_{M_t}^{(k)}$ denotes the $k$-th maximum element in $A_{M_t}$ from timestep $t$. $\mathbb{T}$ is a set of $t$, and has $m = \left | \mathbb{T} \right |$. With $N$ concepts, we normalize the highest activation score $S_i$ of each $\rm [V_i]$ by $\Bar{S}_i = {S_i} / \sum_{j=1}^N S_j$, and apply a \textit{Softmax} operation to $\Bar{S}_i$ to obtain the sampling ratio $r_i$:

\begin{equation}
\label{eq:9}
    r_i = 1 - \frac{e^{\Bar{S}_i}}{\sum_{j=1}^N e^{\Bar{S}_j}}
\end{equation}

Since the initialization of $\rm [V]$ induces differences in attention activation and requires varying training steps for each concept, the proposed adaptive sampling ratio, $r_i$, can appropriately adjust the sampling frequency based on activation scores, thereby improving synchronicity.

\subsection{Feature-retaining training framework}
\label{sec3.4}
The goal of multiple-concept disentanglement is to learn multiple concepts from a single image and sample individual concepts or concept groups with minimal distortion. By applying a mask for each concept, multiple concepts can be disentangled through  the combination of a masked reconstruction loss and a cross-attention loss, expressed as:

\begin{equation}
\label{eq:10}
    L_{rec} = \mathbb{E}_{z,y_s,t, \varepsilon} \left [ \left \| 
    \left ( \varepsilon - \varepsilon_{\theta} \left ( z_t,t,\tau_\theta(y_s) \right ) \right ) \odot M_s 
    \right \|^2_2 \right ]
\end{equation}
\begin{equation}
\label{eq:11}
    L_{attn} = \frac{1}{ \left | s \right |} \sum_{i \in s} \left \| A_C \left ( v_i, z_t \right ) - M_i \right \|_2^2
\end{equation}

\noindent where $y_s$ and $M_s$ are text prompts and masks for a sampled subset $s$, and $A_C \left ( v_i, z_t \right )$ denotes the cross-attention map between $\rm [V_i]$ in $s$ and $z_t$. $M_i \in M_s$ is the corresponding mask. $L_{rec}$ promotes the model to learn features of target concepts, and $L_{attn}$ helps disentangle concepts \cite{avrahami2023break}.

Nevertheless, when $\left | s \right | > 1$, the back-propagation of $L_{rec}$, which contains features from multiple concepts, may induce feature fusion. Therefore, we propose an optimized feature-retaining framework for multiple-concept disentanglement by introducing different training objectives for $s$ of varying sizes. Concretely, for $\left | s \right | = 1$, both $L_{rec}$ and $L_{attn}$ are back-propagated to jointly learn visual features and establish explicit connections between $\rm [V]$ and the visual features. In contrast, only $L_{attn}$ is back-propagated when $\left | s \right | > 1$, preventing feature fusion and enforcing the model to disentangle the cross-attention of multiple concepts while learning their joint presence. The general loss function is formulated as follows:

\begin{equation}
\label{eq:12}
    L  = \left\{
    \begin{matrix}
    L_{rec} + \alpha L_{attn} & \textit{if} \left | s \right | = 1 \\
    L_{attn}  & \textit{if} \left | s \right | > 1
    \end{matrix}
    \right.
\end{equation}

\noindent where $\alpha$ is a scaling coefficient. Moreover, we design a hyperparameter $\omega$ as the proportion of multiple-concept sampling steps (\ie $\left | s \right | > 1$), and apply the adaptive sampling ratio of each concept (see \cref{sec3.2}) when $\left | s \right | = 1$ to facilitate synchronized feature learning across different concepts. This forms the training pipeline of \textit{AttenCraft}, as shown at the bottom of \cref{fig:2}.

\section{Experiments}
\label{4.experiments}

\begin{figure*}[htb!]
  \centering
    \includegraphics[width=1.0\linewidth]{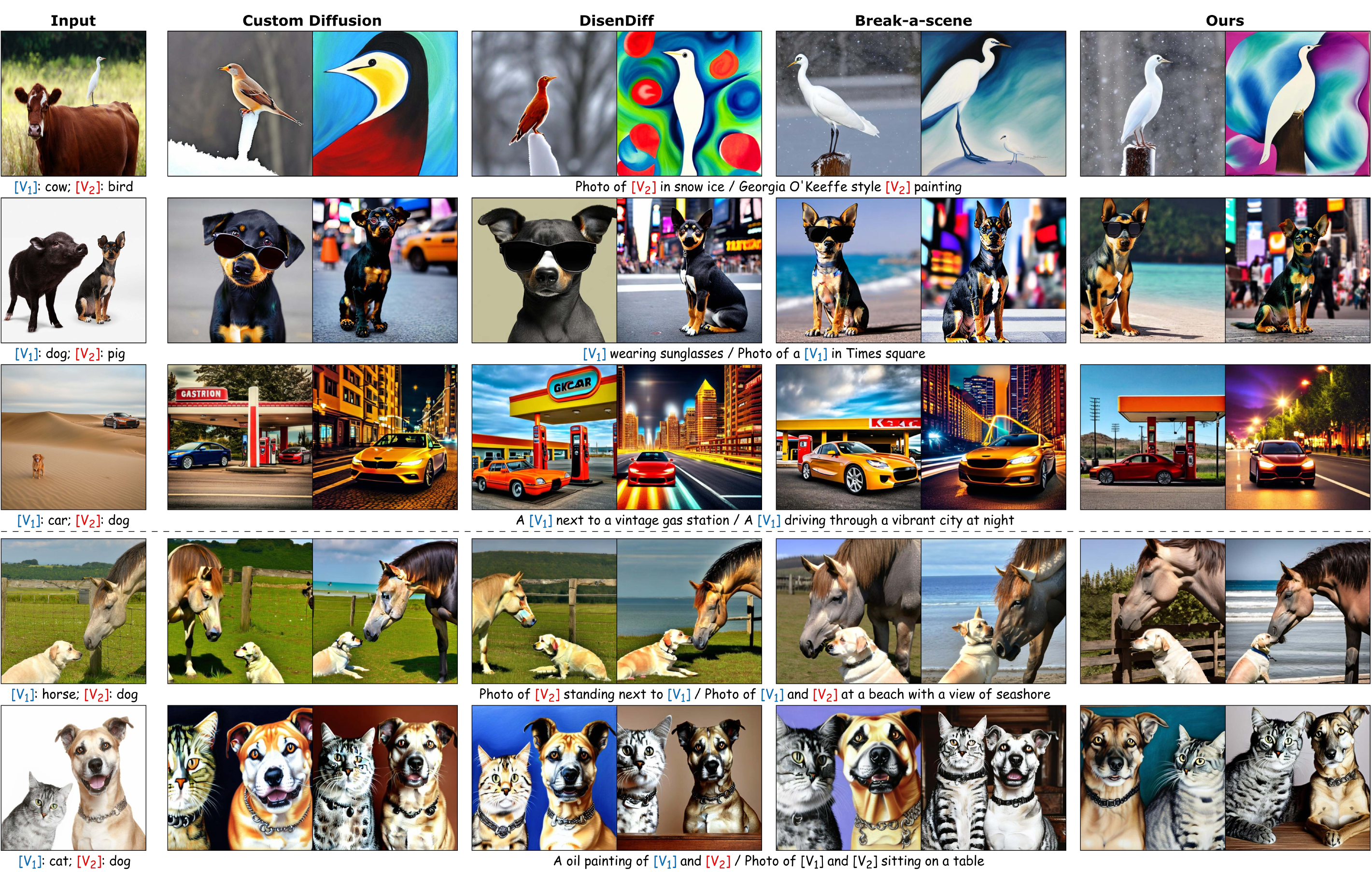}

   \caption{\textbf{Qualitative results for concept disentanglement and feature fusion.} \textit{CusDiff} cannot disentangle multiple concepts, and both \textit{DisenDiff} and \textit{BAS} present feature fusion. Our method not only disentangles the target concepts, but also mitigates the feature fusion problems .}
   \label{fig:5}
\end{figure*}

\subsection{Experimental settings}
\label{sec 4.1}
\paragraph{Datasets and baseline.} We conduct experiments on 16 datasets across various categories, including human, animal, and object. We collect 10 datasets with relatively simple backgrounds from \textit{DisenDiff} \cite{zhang2024attention}. We also synthesize 6 datasets using \textit{Gen4Gen} \cite{yeh2024gen4gen}, which combines multiple personalized concepts into complex backgrounds sourced from copyright-free platforms \footnote{\url{https://unsplash.com}}, where the concepts are curated from the \textit{DreamBooth} \cite{ruiz2023dreambooth} and \textit{CustomConcept101} \cite{kumari2023multi}. We compare our method with \textit{BAS} \cite{avrahami2023break} and \textit{DisenDiff} \cite{zhang2024attention}. Additionally, we implement \textit{CusDiff} \cite{kumari2023multi} to demonstrate the disentanglement capability of general subject-driven models.

\paragraph{Evaluation metrics.} Following baseline models, we calculate the CLIP-I and CLIP-T scores to assess image fidelity and prompt fidelity. Specifically, CLIP-I represents the cosine similarity between the CLIP-ViT-L/14 embeddings of generated and input images, while CLIP-T measures that between generated images and text prompts. In addition, we calculate the DINO score, which is the cosine similarity between the ViT-B/16 DINO-V2 embeddings of the generated and input images, to reveal how much the model preserves the concept identity. Depending on the dataset and concept evaluation scope, CLIP-I and DINO scores use different target references: \textit{DisenDiff} datasets uses cropped input images for concept subsets and the original input for all concepts, while \textit{Gen4Gen} datasets uses original single-concept images for subsets and the generated composite input image for all concepts. Moreover, we evaluate the learning synchronicity using CLIP-I-sync, which is the absolute difference between CLIP-I scores of single concepts in the same dataset. For each dataset, we prepare 10 text prompts for single concepts and the concept group, respectively. We generate 10 images for each text prompt using 50 steps of the PNDM scheduler \cite{liu2022pseudo} with a guidance scale of 7.5, resulting in an evaluation set consisting of 300 images.

\paragraph{Implementation details.} We use SD v2.1 trained on the LAION-5B dataset \cite{schuhmann2022laion} as the base model. We initialize each identifier token $\rm [V]$ with the text embedding of the corresponding class name. We extract cross-attention and self-attention maps from the attention layers, with dimensions of $16 \times 16$ and $32 \times 32$, respectively. These maps contain abundant semantic and visual information \cite{hertz2022prompt, nguyen2024dataset}. We set the powers $\upsilon$ and $\tau$ to 2 and 4, respectively \cite{zhang2024attention, nguyen2024dataset}. The threshold $\gamma$ for \textit{Delta masking} is empirically set to 0.1. An illustration of the initialized masks is provided in the supplement. When extracting attention scores as described in \cref{eq:7}, we set $n=5$ and $\mathbb{T} = \left \{ 0, 20, 40, 60, 80 \right \}$. Moreover, we set the scaling coefficient $\alpha = 0.01$, and the ratio $\omega = 0.3$. All experiments are conducted on an NVIDIA A100 GPU with a single input image, a batch size of 1, and a learning rate of $1 \times 10^{-4}$ for 300 steps. To reduce computational costs, only the $W_k$ and $W_v$ matrices in the cross-attention layers are trained \cite{kumari2023multi}. Implementation details of baseline models are provided in the supplement.

\subsection{Qualitative comparisons}
\label{4.2}

\begin{figure*}[htb!]
  \centering
    \includegraphics[width=0.9\linewidth]{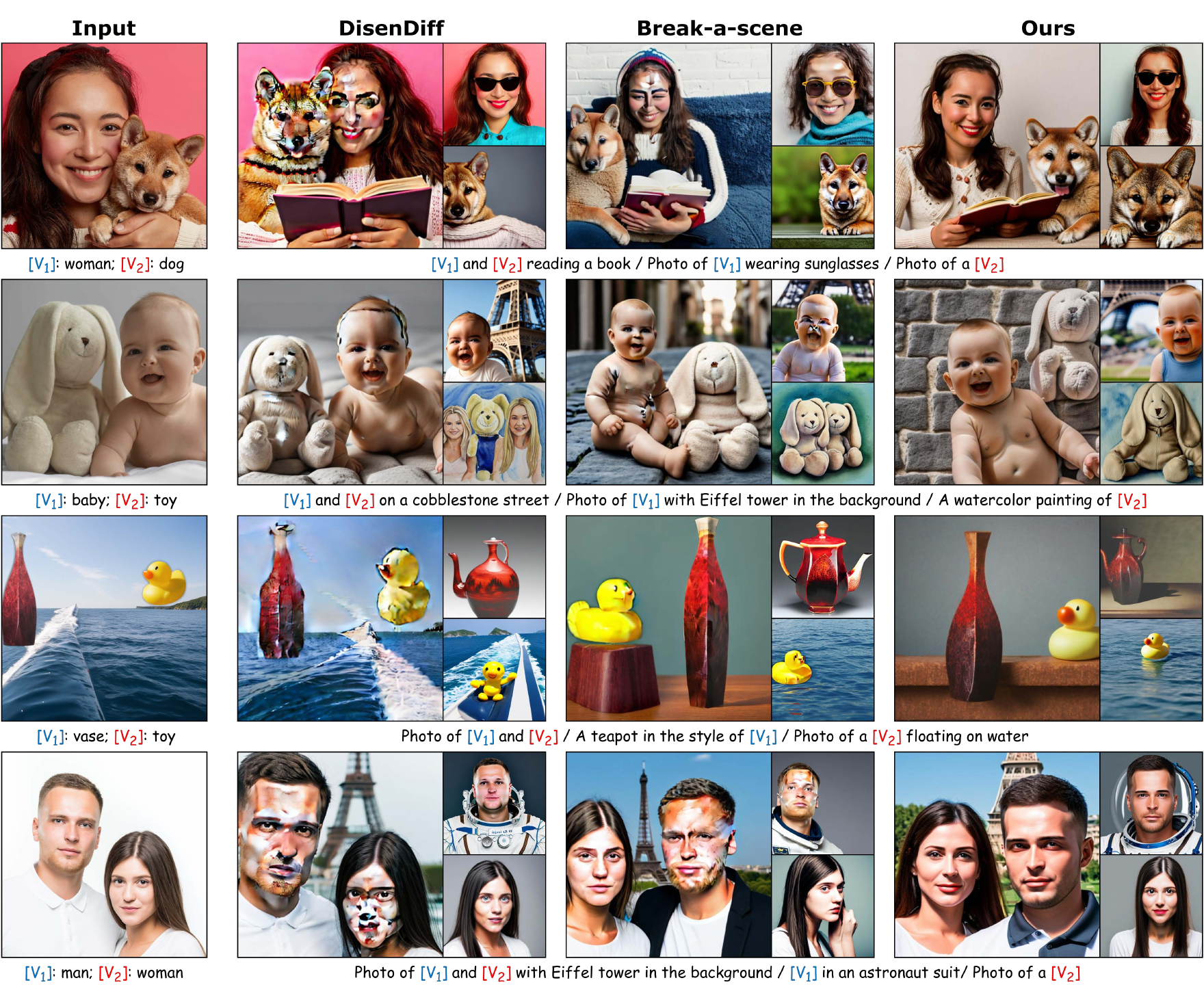}

   \caption{\textbf{Qualitative results for learning synchronicity.} \textit{DisenDiff} and \textit{BAS} show asynchronous learning in different forms, while our method achieves a more synchronous feature learning.}
   \label{fig:10}
\end{figure*}

We present a qualitative comparison between our method and baseline models in \cref{fig:5} and \cref{fig:10}. \Cref{fig:5} presents the generated images of single concepts and concept groups from different datasets to illustrate the models' performance in disentangling multiple concepts, and the presence of feature fusion. Upon examination, \textit{CusDiff} struggles to disentangle multiple concepts, as the generated images for single concepts show distinct features from the input images. The concept group images generated by \textit{CusDiff} present all target concepts, but the feature fusion can be spotted from the color of the concepts. On the other hand, \textit{DisenDiff} and \textit{BAS} present disentangling capability, but the problems of feature fusion still stand out. \textit{DisenDiff} shows blended features in both single concept and concept group images (\eg, color of the bird, dog, and car for single concept; necklace of the cat and dog, color of the horse and dog for concept group). Also, both \textit{CusDiff} and \textit{DisenDiff} present background features in the ``horse \& dog'' dataset, as the grass appears with the target concepts, indicating that the model fails to detach background features from the target concept. While \textit{BAS} exhibits fewer blended features than \textit{DisenDiff}, the feature fusion can still be observed (\eg, color of the car for the single concept; ear shape of the dog, necklace and color of the cat and dog for concept group). In contrast, our method shows clear disentanglement across multiple concepts and background information, and the features from each target concept are well-retained without blending and fusion.

Furthermore, we analyze the learning synchronicity across multiple concepts of models capable of disentangling and learning them separately. \Cref{fig:10} presents examples of image triplets consisting of single concept and concept group images generated by the model undergoing the same training step for a fair comparison of learning synchronicity. Different forms of asynchronous learning can be observed from \textit{DisenDiff} and \textit{BAS}. Specifically, \textit{DisenDiff} tends to show asynchronous learning between single concepts and concept groups, and overfits the concept group before single concepts (manifest by the corruption in specific regions). In the ``baby \& toy'' and ``toy \& vase'' datasets, the concept groups are overfit while the target toys are not fully learned. On the other hand, \textit{BAS} usually presents asynchronous learning across single concepts, and overfits one of the target concepts. Our method exhibits better learning synchronicity compared to the baseline models, as reflected by the results.

\begin{figure}[htb!]
  \centering
    \includegraphics[width=0.95\linewidth]{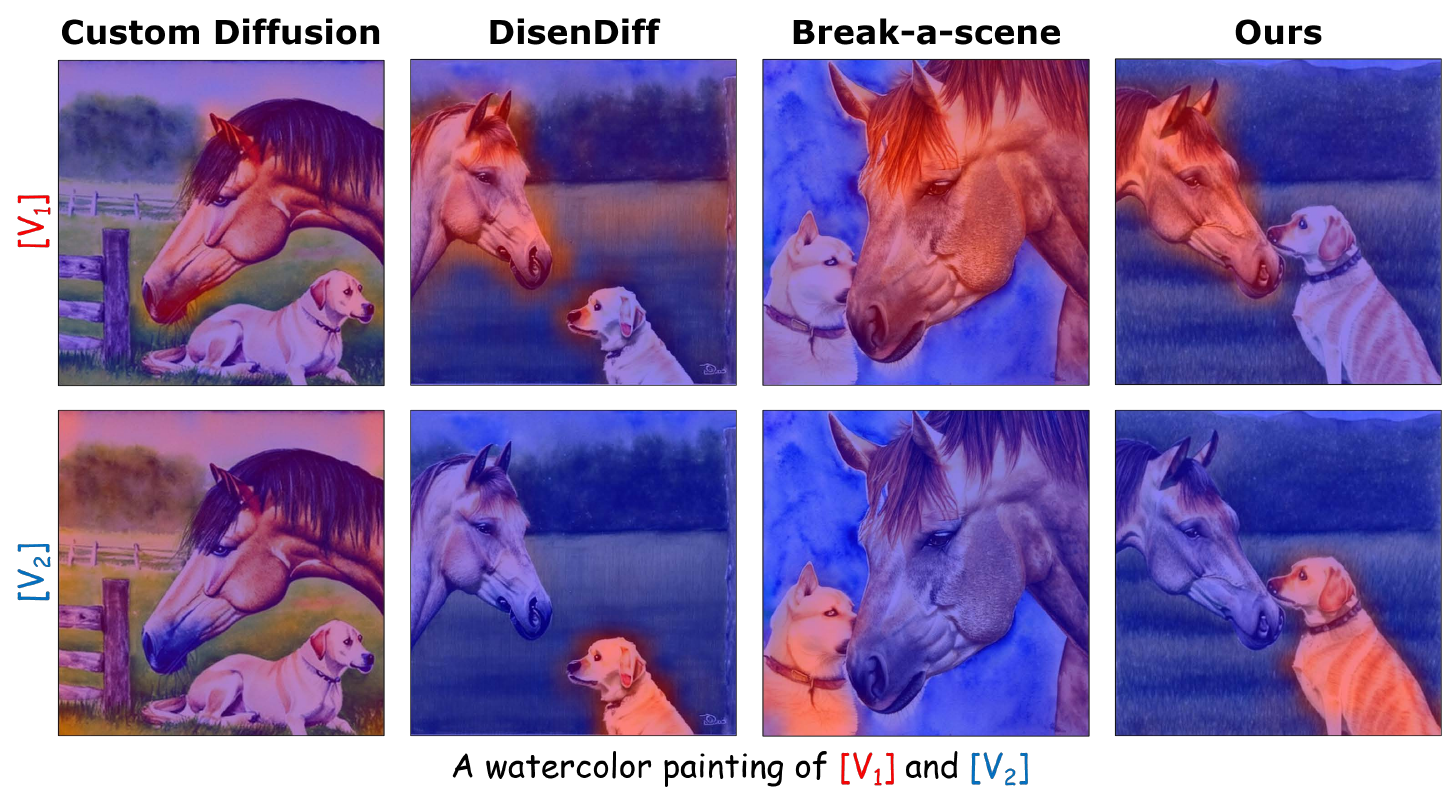}
   \caption{\textbf{Visualization of cross-attention maps.} Our method presents proper attention activation for multiple conceptions.}
   \label{fig:6}
\end{figure}

In addition, we assess the models' disentanglement capabilities by visualizing the cross-attention maps of each $\rm [V]$. As demonstrated in \cref{fig:6}, although all models generate both target concepts, \textit{CusDiff} fails to show appropriate attention activation fitting the concepts, indicating that it does not disentangle them. Moreover, \textit{DisenDiff} displays attention activation on the background for $\rm [V_1]$ apart from the horse, suggesting that it struggles to eliminate the background. While \textit{BAS} demonstrates attention activation consistent with the concepts,  it fails to accurately depict the dog's appearance as pointy ears are observed. Our model shows strong consistency between the attention maps and concepts, effectively highlighting the concept features.

\subsection{Quantitative comparisons}
\label{sec 4.3}

Quantitative comparisons between our method and baseline models are presented in \cref{table:1}. Despite \textit{CusDiff} achieving the second-highest score in concept-group CLIP-I, it has been shown to be incapable of disentangling multiple concepts, resulting in the lowest single-concept CLIP-I score. \textit{BAS} displays higher CLIP-I scores but lower CLIP-T and group DINO scores compared to \textit{DisenDiff}. Notably, our method surpasses all baseline models in the CLIP-I and DINO scores across both scenarios. Regarding the CLIP-T score, our method ranks second in concept-group generation with only marginal differences compared to baseline models. Remarkably, our method records the lowest CLIP-I-sync score, demonstrating its improvement in learning synchronicity. Although \textit{CusDiff} has the second-lowest CLIP-I-sync score, the high synchronicity stems from its failure to learn single concepts effectively. \textit{BAS}, benefiting from the union sampling scheme, achieves a better CLIP-I-sync score than \textit{DisenDiff}.

\begin{table}[htb]
  \centering
    \caption{ \raggedright Results of quantitative comparisons}
\small{
  \begin{tabular}{@{}lcccc@{}}
    \toprule
    Metrics     & CusDiff & DisenDiff & BAS & Ours \\
    \midrule
    \multicolumn{5}{@{}l}{\textit{Single concept}} \\
    CLIP-I$\uparrow$ & 0.531 & 0.554 & \underline{0.563} & \textbf{0.576} \\
    CLIP-T$\uparrow$ & \textbf{0.186} & \underline{0.185} & 0.178 & 0.184 \\
    DINO$\uparrow$   & 0.653    & 0.657   &  \underline{0.666}  &  \textbf{0.703}   \\
    \midrule
    \multicolumn{5}{@{}l}{\textit{Concept group}} \\
    CLIP-I$\uparrow$ & \underline{0.567} & 0.553 & 0.557 & \textbf{0.584} \\
    CLIP-T$\uparrow$ & \textbf{0.213} & 0.202 & 0.201 & \underline{0.209} \\
    DINO$\uparrow$   & 0.654    & \underline{0.684}  &  0.654  &  \textbf{0.723}   \\
    \midrule
    CLIP-I-sync$\downarrow$ & \underline{0.089} & 0.097 & 0.091 & \textbf{0.062}  \\
    \bottomrule
  \end{tabular}
  }
  \label{table:1}
\end{table}

\subsection{Ablation studies}
\label{sec 4.4}

\begin{figure}[htb!]
  \centering
    \includegraphics[width=0.95\linewidth]{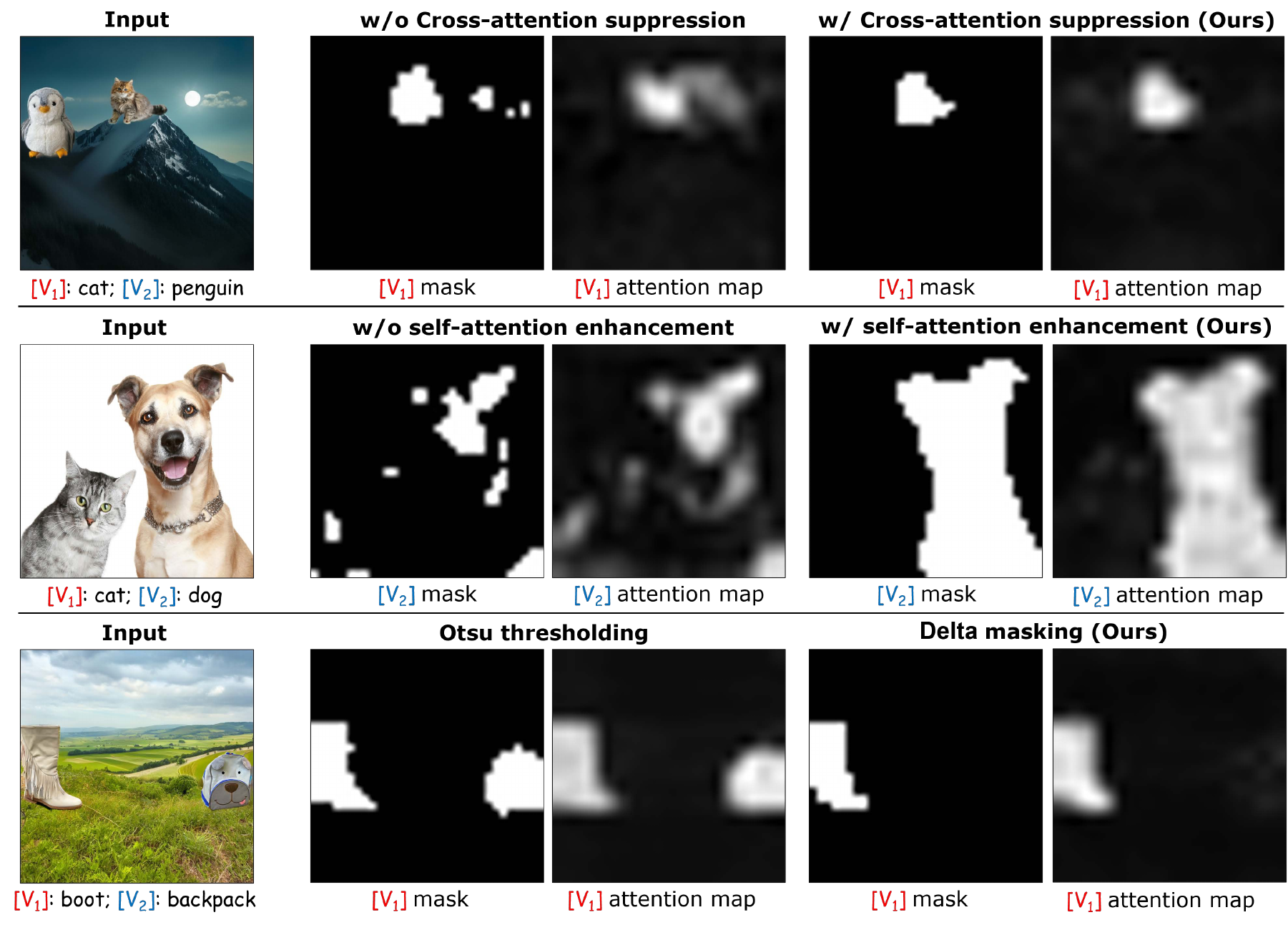}
   \caption{\textbf{Qualitative results for ablating attention-guided mask creation.} All three techniques are vital for mask creation, and disabling them will cause failure in certain datasets.}
   \label{fig:7}
\end{figure}

\paragraph{Attention-guided mask creation} 
First, we ablate the attention-guided mask creation process by individually disabling each of the three key techniques. We find that disabling \textit{Cross-attention suppression} permits weak attention activations outside the concept region, resulting in fragmented mask activations. Moreover, omitting \textit{Self-attention enhancement} results in uneven and unsmooth attention distributions within the target region, producing low-quality masks. Furthermore, we substitute \textit{Delta masking} with Otsu thresholding and observe that the latter often fails to separate masks of different concepts, leading to incorrect associations between identifier tokens and corresponding visual features. Representative cases are presented in \cref{fig:7}. Therefore, combining the three key techniques ensures the creation of high-quality masks, which guide the precise disentanglement of multiple concepts.

\begin{figure}[htb!]
  \centering
    \includegraphics[width=0.95\linewidth]{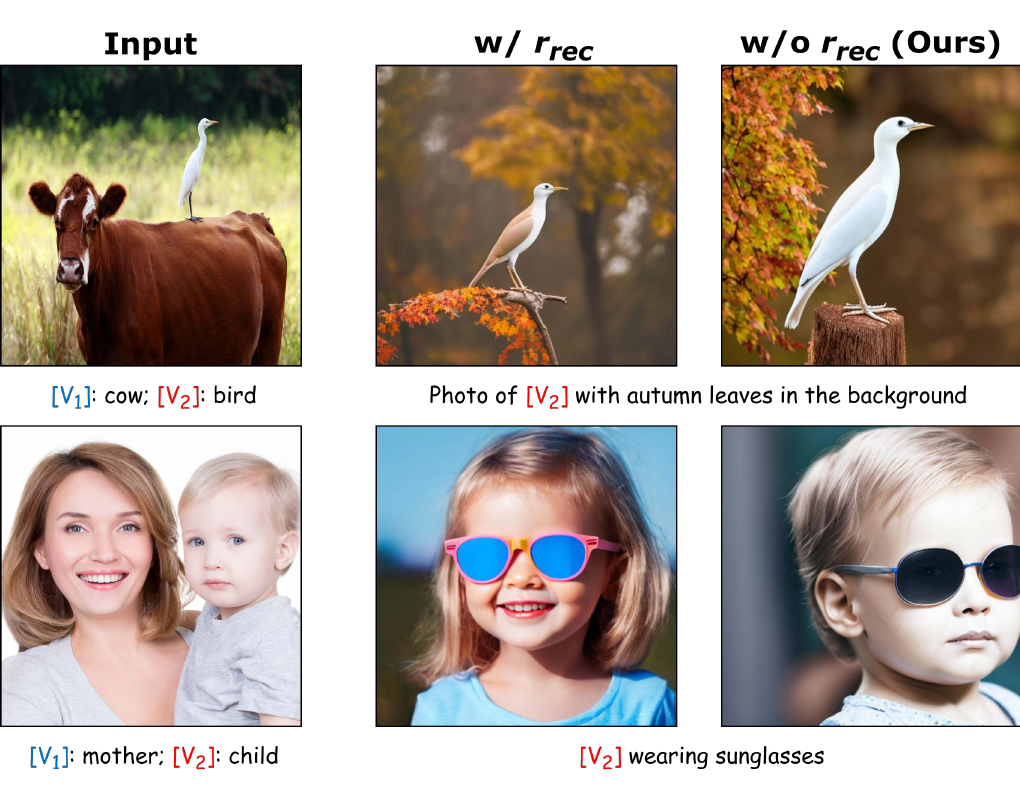}
   \caption{\textbf{Qualitative results of ablation studies on feature-retaining training framework.} Our proposed framework can effectively prevent feature fusion during training.}
   \label{app:fig8}
\end{figure}

\begin{figure}
  \centering
  \begin{subfigure}{0.47\linewidth} 
    \includegraphics[width=1.0\linewidth]{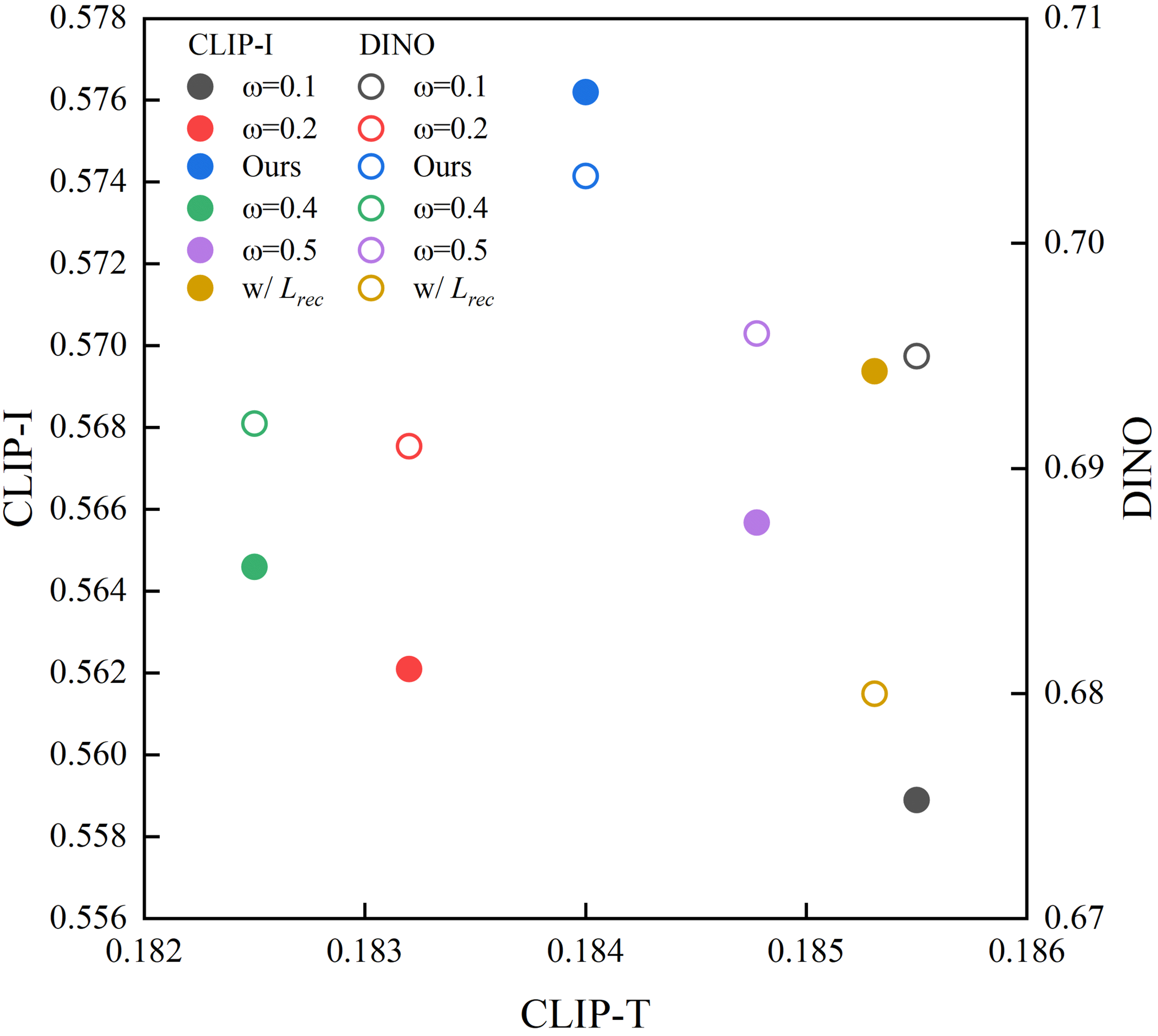}
    \caption{}
    \label{fig:9a}
  \end{subfigure}
  \hfill
  \begin{subfigure}{0.47\linewidth} 
    \includegraphics[width=1.0\linewidth]{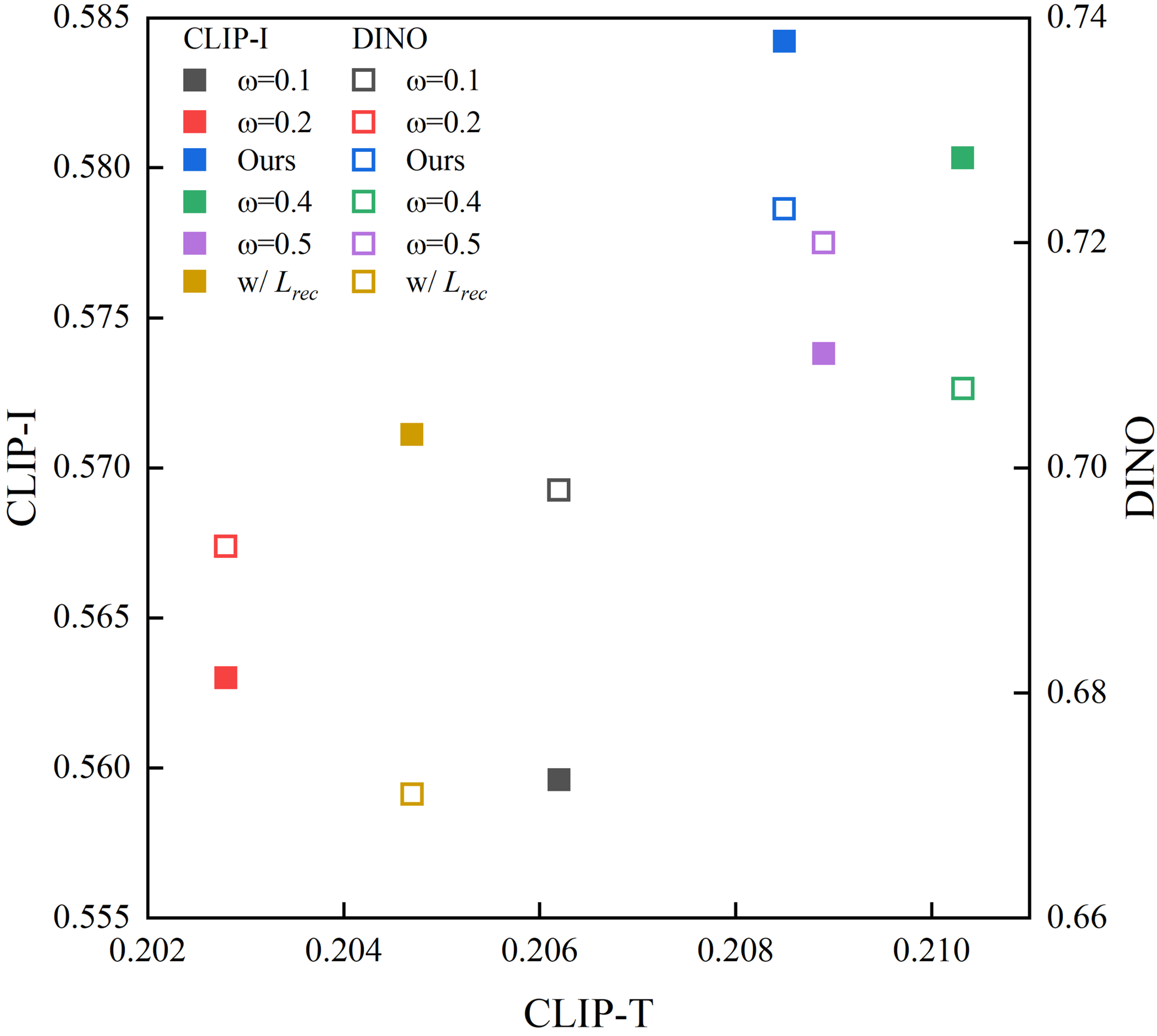}
    \caption{}
    \label{fig:9b}
  \end{subfigure}
  \caption{\textbf{Quantitative results of ablating feature-retaining training framework.} Our method shows the best overall performance concerning (a) single concept and (b) concept group.}
  \label{fig:9}
\end{figure}

\paragraph{Feature-retaining training framework}
We propose the feature-retaining training framework by applying different loss functions to sampled subsets $s$ of varying sizes. To validate the optimized framework, we compare our method with a variant that back-propagates $L_{rec}$ when $\left | s \right | > 1$. A quantitative result comparison is presented in \cref{app:fig8}, revealing that back-propagating $L_{rec}$ can increase the risk of feature fusion, as evidenced by the color of bird and the hairstyle of baby. Quantitative results in \cref{fig:9} indicate that back-propagating $L_{rec}$ reduces CLIP-I and DINO scores while slightly increasing single-concept CLIP-T scores. In addition,  we investigate the value of $\omega$, as over-sampling single concepts would impair the model's ability to generate multiple concepts, whereas over-sampling multiple concepts would delay feature learning. Model performance with $\omega$ ranging from 0.1 to 0.5 is illustrated in \cref{fig:9}. Our method with $\omega=0.3$ achieves the highest CLIP-I and DINO scores, with only a marginal difference in CLIP-T score.

\begin{figure}[htb!]
  \centering
    \includegraphics[width=0.9\linewidth]{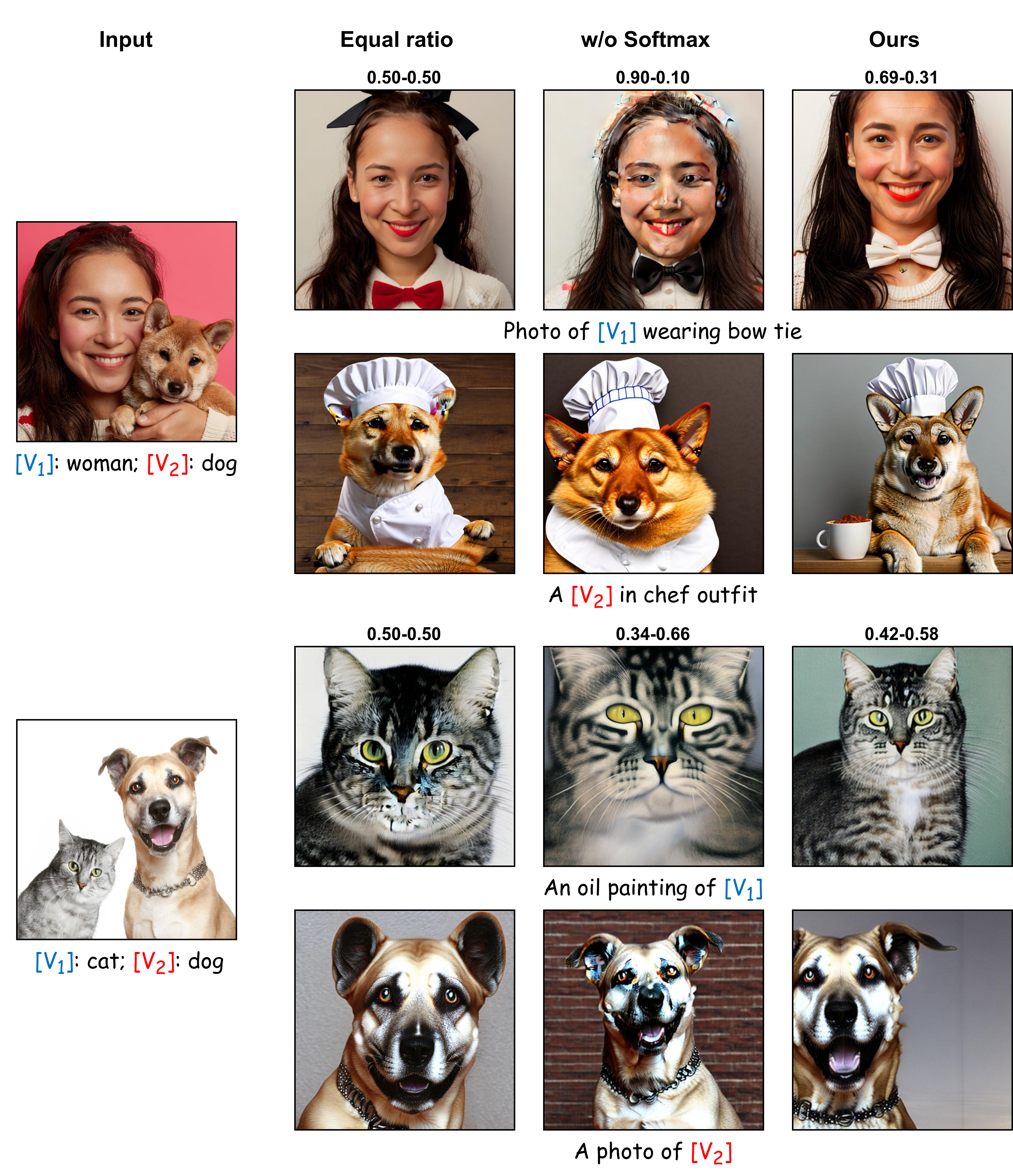}
   \caption{\textbf{Qualitative results of ablation studies on adaptive estimation of sampling ratio.} The numbers on images denote the sampling ratio determined by the method.}
   \label{app:fig9}
\end{figure}

\paragraph{Adaptive sampling ratio estimation}
As shown in \cref{eq:9}, the adaptive sampling ratio is estimated using attention activation scores through normalization and \textit{Softmax} operations. Thus, we evaluate the model performance under two modifications: (1) applying an equal sampling ratio (0.5-0.5) to single concepts and (2) disabling the \textit{Softmax} to validate our design. We select six datasets in which the difference in the estimated sampling ratio between the two concepts exceeds 0.5 to allow for an apparent comparison. \Cref{table:2} lists the evaluation results, which indicate that using an equal sampling ratio or disabling the \textit{Softmax} operation degrades the fidelity of generated images and increases the disparity in learning synchronicity across different concepts. A qualitative comparison is presented in \cref{app:fig9}.

\begin{table}[htb]
  \centering
    \caption{ \raggedright Ablation results of adaptive sampling ratio estimation}
    \small{
  \begin{tabular}{@{}lccc@{}}
    \toprule
    Metrics     & Equal ratio & w/o \textit{softmax} & Ours \\
    \midrule
    \multicolumn{4}{@{}l}{\textit{Single concept}} \\
    CLIP-I$\uparrow$ & 0.580 & \underline{0.581} & \textbf{0.585} \\
    CLIP-T$\uparrow$ & \underline{0.182} & \textbf{0.183} & \underline{0.182} \\
    DINO$\uparrow$ & 0.673 & \underline{0.680} & \textbf{0.688} \\
    \midrule
    \multicolumn{4}{@{}l}{\textit{Concept group}} \\
    CLIP-I$\uparrow$ & \underline{0.581} & 0.578 & \textbf{0.607} \\
    CLIP-T$\uparrow$ & 0.207 & \underline{0.209} & \textbf{0.211} \\
    DINO$\uparrow$ & \underline{0.715} & 0.704 & \textbf{0.723} \\
    \midrule
    CLIP-I-sync$\downarrow$ & \underline{0.046} & 0.053 & \textbf{0.041}  \\
    \bottomrule
  \end{tabular}
  }
  \label{table:2}
\end{table}

\subsection{Generalizing to more concepts}
While our main experiments focus on two-concept cases for fair comparison with baselines, the proposed innovations (i.e., attention-guided mask generation, adaptive sampling ratio estimation, and feature-retaining training) in AttenCraft are inherently scalable to images with more than two concepts. Each module can operate on arbitrary numbers of concepts (via multi-mask generation, normalized attention-based ratios, and concept-specific loss design), making the method directly applicable beyond two-concept scenarios. 

\begin{figure}[htb]
  \centering
   \includegraphics[width=1.0\linewidth]{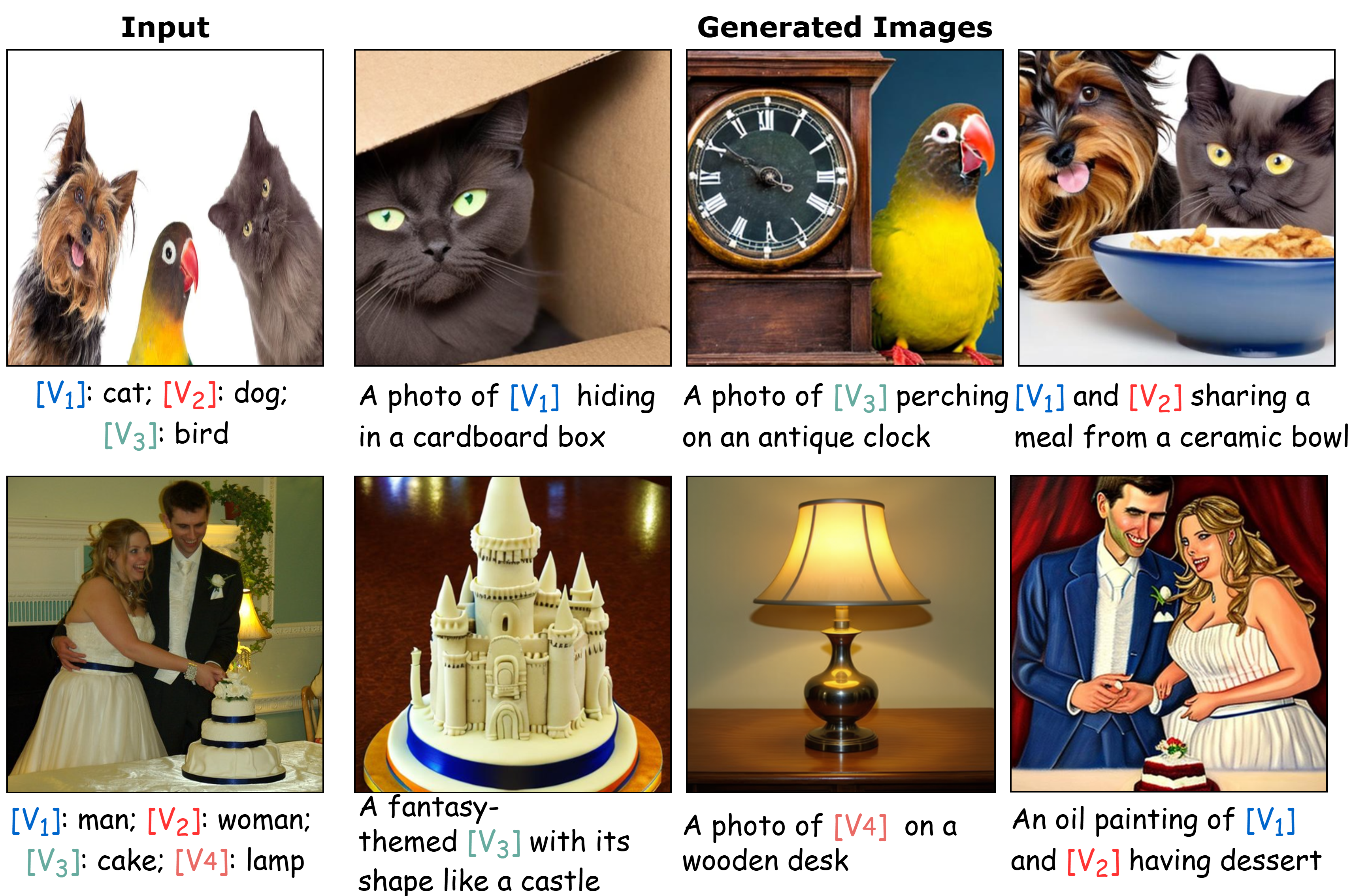}

   \caption{\textbf{Qualitative results for \textit{AttenCraft} applied on input images containing more than two concepts.} Our proposed method can be seamlessly applied to input images containing more than two concepts.}
   \label{revise:1}
\end{figure}

To further support our claim, we conduct additional experiments using images that contain more than two concepts as inputs to our method. The generated results are shown in \cref{revise:1}. As observed, even with more complex inputs, our method successfully disentangles the concepts and produces coherent generations for both single concepts and concept groups. It should be noted that, given that the lamp's base in the second row was occluded in the input image, our method is nonetheless able to synthesize a new base. The lamp shade, however, is learned precisely and is rendered accurately.

\section{Conclusion}
In this paper, we identify two key issues in diffusion-based T2I models designed to disentangle multiple concepts from a single input image for T2I customization: feature fusion and asynchronous learning. To mitigate them, we propose a novel attention-based method named \textit{AttenCraft} as an optimized solution. We investigate the relationship between feature acquisition and identifier token initialization, and introduce an adaptive algorithm based on cross-attention scores for automatically estimating the sampling ratio of multiple concepts to mitigate asynchronous learning. Moreover, we optimize the training framework by introducing different loss functions for sampled subsets of varying sizes, retaining concept features and preventing feature fusion. In addition, we utilize attention maps to create accurate masks for each concept to guide disentanglement within a single step, without using specialized models or human inputs.

\bibliographystyle{IEEEtran}
\bibliography{main.bib}

\vfill

\clearpage

\appendix

\section*{Datasets}
\label{app:datasets}

\begin{figure}[htb!]
    \centering
    \includegraphics[width=0.9\linewidth]{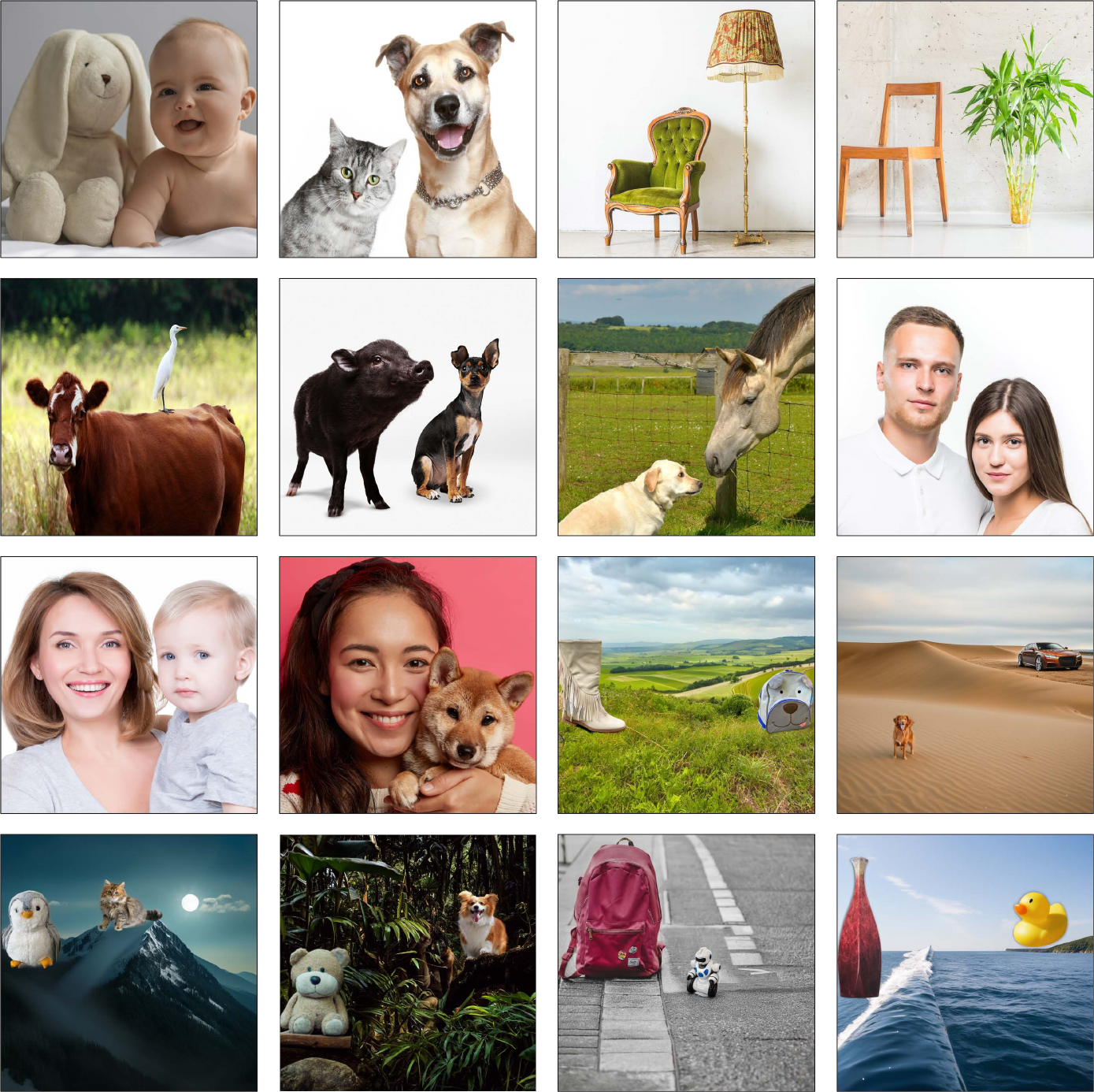}
    \caption{Illustration of datasets}
    \label{app:fig1}
\end{figure}

In this study, we curate 16 datasets for experiment and evaluation. We include 10 datasets introduced by \textit{DisenDiff} \cite{zhang2024attention}, generally featuring simple backgrounds. Additionally, we utilized the Gen4Gen dataset creation pipeline \cite{yeh2024gen4gen} to amalgamate personalized concepts into complex backgrounds (\eg fields, mountains, forests) sourced from copyright-free platforms, resulting in 6 synthetic datasets. The personalized concepts used for dataset synthesis were collected from the \textit{DreamBooth} dataset \cite{ruiz2023dreambooth} and \textit{CustomConcept101} \cite{kumari2023multi}. All datasets are presented in \cref{app:fig1}, and the class names for each dataset are: (1) baby \& toy; (2) cat \& dog; (3) chair \& lamp; (4) chair \& vase; (5) cow \& bird; (6) dog \& pig; (7) horse \& dog;  (8) man \& woman; (9) mother \& child; (10) woman \& dog; (11) boot \& backpack; (12) car \& dog; (13) cat \& penguin; (14) dog \& bear; (15) backpack \& toy; (16) vase \& toy.

\section*{Additional details for preliminary experiment}
\label{app:pre-experiments}
We introduce a preliminary experiment in \cref{sec3.3} to evaluate how the initialization of the identifier token $\rm [V]$ affects feature acquisition. The experiment utilizes the dataset (1)-(10) introduced in \cref{app:datasets}, employing \textit{BAS} to disentangle multiple concepts and learn each one individually. The learning rate is set to $5 \times 10^{-5}$, the maximum training steps to 1000, and the first textual inversion phase is disabled to improve efficiency. We design three initialization patterns for the token, consisting of combinations of the text embeddings of the precise class (dubbed as \textit{P}) and the general category (dubbed as \textit{G}) of the target concept. For each dataset, two target concepts are initialized by a triplet of \textit{P}-\textit{P}, \textit{P}-\textit{G}, and \textit{G}-\textit{P}, respectively. The complete list of triplets for all datasets is provided in \cref{app:table1}.

\begin{table}[htb!]
  \centering
    \caption{ \raggedright Patterns for initialization of identifier tokens}
    \small{
  \begin{tabular}{@{}lccc@{}}
    \toprule
    Dataset     & \textit{P-P} & \textit{P-G} & \textit{G-P} \\
    \midrule
    baby \& toy & baby-rabbit & baby-toy & human-rabbit \\
    cat \& dog  & cat-dog & cat-animal & animal-dog \\
    chair \& lamp & chair-lamp & chair-lighting & furniture-lamp \\
    chair \& vase & chair-vase & chair-decor & furniture-vase \\
    cow \& bird & cow-bird & cow-animal & animal-bird \\
    dog \& pig & dog-pig & dog-animal & animal-pig \\
    horse \& dog & horse-dog & horse-animal & animal-dog \\
    man \& woman & man-woman & man-human & human-woman \\
    mother \& child & mother-child & mother-human & human-child \\
    woman \& dog & woman-dog & woman-animal & human-dog \\
    \bottomrule
  \end{tabular}
  }
  \label{app:table1}
\end{table}

\begin{table}[htb!]
  \centering
    \caption{ \raggedright Highest cross-attention scores of $\rm [V]$ using different initialization patterns}
    \small{
  \begin{tabular}{@{}lcccc@{}}
    \toprule
    Dataset    & Concept & \textit{P-P} & \textit{P-G} & \textit{G-P} \\
    \midrule
    \multirow{2}{*}{baby \& toy} & baby    &  0.020   &  0.020   &  0.004   \\
                               & toy     &  0.026   &  0.015   &  0.026   \\
    \hline
    \multirow{2}{*}{cat \& dog}  & cat     &  0.039   &  0.042   &  0.010   \\ 
                               & dog     &  0.020   &  0.005   &  0.023   \\
    \hline
    \multirow{2}{*}{chair \& lamp}  & chair     &  0.011   &  0.011   &  0.005  \\
                                  & lamp      &  0.012   &  0.006   &  0.014   \\
    \hline
    \multirow{2}{*}{chair \& vase}  & chair     &  0.013   &  0.014   &  0.006   \\
                                  & vase      &  0.011   &  0.005   &  0.012   \\
    \hline
    \multirow{2}{*}{cow \& bird}  & cow     &  0.046   &  0.045   &  0.010   \\
                               & bird     &  0.027   &  0.005   &  0.014   \\
    \hline
    \multirow{2}{*}{dog \& pig}  & dog     &  0.036   &  0.040   &  0.013   \\
                               & pig     &  0.052   &  0.005   &  0.062   \\
    \hline
    \multirow{2}{*}{horse \& dog}  & horse     &  0.031   &  0.032   &  0.008   \\ 
                                 & dog       &  0.031   &  0.009   &  0.032   \\
    \hline
    \multirow{2}{*}{man \& woman}  & man     &  0.003   &  0.003   &  0.003   \\
                                 & woman     &  0.003   &  0.002   &  0.004   \\
    \hline
    \multirow{2}{*}{mother \& child}  & mother      &  0.007   &  0.010   &  0.002   \\ 
                                    & child       &  0.012   &  0.003   &  0.012   \\
    \hline
    \multirow{2}{*}{woman \& dog}  & woman     &  0.005   &  0.004   &  0.003   \\
                                 & dog       &  0.043   &  0.023   &  0.036   \\
    \bottomrule
  \end{tabular}
  }
  \label{app:table2}
\end{table}

\begin{figure*}[htb!]
    \centering
    \includegraphics[width=0.9\linewidth]{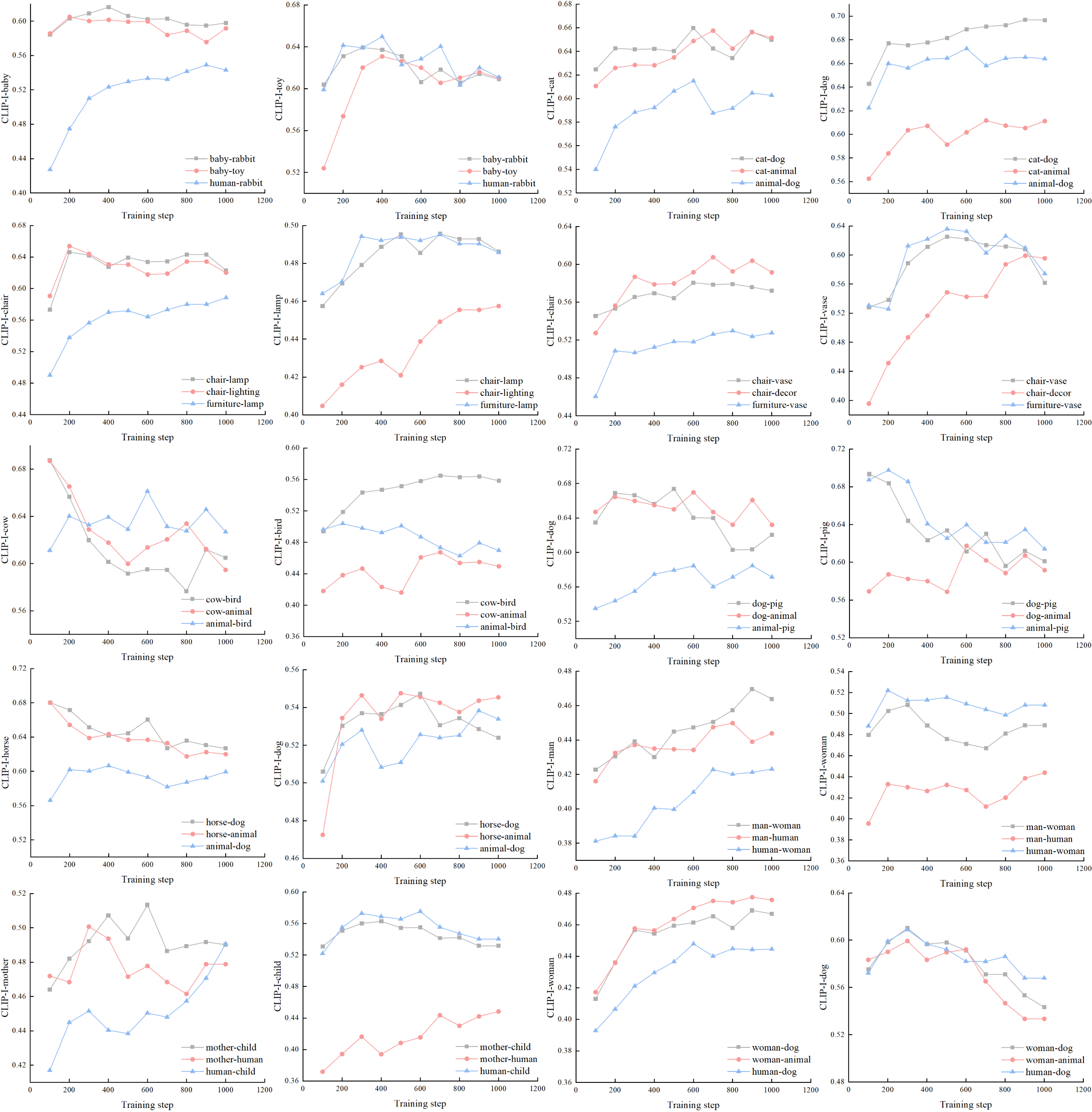}
    \caption{Variation of single concept CLIP-I scores with training step}
    \label{app:fig2}
\end{figure*}

We evaluate the single-concept CLIP-I scores of images generated by \textit{BAS} at intervals of 100 training steps, following the evaluation pipeline described in \cref{sec 4.1}. Detailed results for all datasets are presented in \cref{app:fig2}. Notably, the initialization pattern varies relative to concepts within the same dataset. For example, we initialize $\rm [V_1]$ and $\rm [V_2]$ in the ``cat \& dog" dataset by ``cat-dog", ``cat-animal", and ``animal-dog", corresponding to \textit{P}-\textit{P}, \textit{P}-\textit{G}, and \textit{G}-\textit{P}, respectively. The initialization `cat-animal' functions as \textit{P}-\textit{G} for assessing the concept ``cat", but serves as \textit{G}-\textit{P} for assessing the concept ``dog", and a similar relationship applies to ``animal-dog". For most target concepts in the datasets, the CLIP-I score starts higher when $\rm [V]$ is initialized with \textit{P} compared to \textit{G}, highlighting the significant impact of the semantic information in $\rm [V]$ on feature acquisition. However, when initialized with \textit{P}, the CLIP-I score tends to decrease with additional training steps, indicating potential overfitting and corruption. In contrast, when initialized with \textit{G}, the CLIP-I score generally increases throughout training. Additionally, for a specific identifier token $\rm [V]$, the initialization of another token in the same dataset has minimal impact on its feature acquisition. The average variation in the CLIP-I score is shown in \cref{fig:3a}.

Furthermore, we analyze the highest cross-attention scores extracted from the cross-attention maps for each $\rm [V]$, as detailed in \cref{app:table2}. The cross-attention activation under different initialization patterns shows similar trends to feature acquisition, with higher scores observed when $\rm [V]$ is initialized using \textit{P} rather than \textit{G}. Similarly, the initialization of other $\rm [V]$ tokens within the same dataset has negligible effects on the cross-attention score. The average cross-attention scores for each initialization pattern are displayed in \cref{fig:3b}.

\section*{Additional details for main experiment}
\label{app:main-experiments}

\subsection*{Implementation details}
\label{app:implement}

\paragraph{Custom Diffusion.} We utilize the official implementation of Custom Diffusion from the HuggingFace platform \cite{von-platen-etal-2022-diffusers} with 200 training steps, a batch size of 1, and a learning rate of $5 \times 10^{-5}$. An AdamW optimizer with $\beta_1 = 0.9$ and $\beta_2 = 0.999$ is applied. During training, the text prompt is ``$[\rm V_1]$ $[\rm Class_1]$ and $[\rm V_2]$ $[\rm Class_2]$" to fit the original design of \textit{CusDiff}. The same prompt design is also employed during inference. The identifier tokens are initialized by rare token embeddings. PEFT is applied in \textit{CusDiff} so that only the $W_k$ and $W_v$ matrices in cross-attention layers of UNet are optimized.

\paragraph{DisenDiff.} We implement DisenDiff based on the official implementation with 250 training steps, a batch size of 1, and a learning rate of $5 \times 10^{-5}$. The optimizer is the AdamW optimizer with $\beta_1 = 0.9$ and $\beta_2 = 0.999$. Similar to \textit{CusDiff}, the design of text prompt ``$[\rm V_1]$ $[\rm Class_1]$ and $[\rm V_2]$ $[\rm Class_2]$" is applied in \textit{DisenDiff} and the identifier tokens are initialized by rare token embeddings. Also, \textit{DisenDiff} follows the selection of trainable parameters of \textit{CusDiff}.

\paragraph{Break-a-scene.} We combine the official implementation of \textit{BAS} with the implementation presented in \textit{Textual Localization} \cite{shentu2024textual}. Since the original implementation of \textit{BAS} optimizes the whole UNet following \textit{DreamBooth} \cite{ruiz2023dreambooth}, while \textit{Textual Localization} presents a similar method with PEFT by only optimizing the $W_k$ and $W_v$ matrices in cross-attention layers of UNet, following \textit{CusDiff}. To ensure a fair comparison, we adapt the implementation of \textit{BAS} with PEFT. We optimize the text embeddings of identifier tokens with a high learning rate of $5 \times 10^{-4}$ for 400 steps in the first training stage, and train the text encoders and $W_k$ and $W_v$ matrices in cross-attention layers with a low learning rate of $5 \times 10^{-5}$ for 200 steps, with a batch size of 1 applied for both stages. An AdamW optimizer with $\beta_1 = 0.9$ and $\beta_2 = 0.999$ is applied for both stages. The masks are created by jointly using the Grounding DINO \cite{liu2023grounding} and SAM \cite{kirillov2023segment}. Moreover, the design of the text prompt is ``$[\rm V_1]$ and $[\rm V_2]$" where the identifier tokens are initialized by corresponding class name embeddings.

\paragraph{AttenCraft (Ours).} We detail the implementation of our method in \cref{sec 4.1}. An AdamW optimizer with $\beta_1 = 0.9$ and $\beta_2 = 0.999$ is applied. 

For completeness, we note that we performed systematic experiments across multiple learning rates ($1 \times 10^{-5}$, $5 \times 10^{-5}$, and $1 \times 10^{-4}$) and a range of fine-tuning steps for each baseline and our method. We then reported the best-performing trial for each method in the main results (\cref{table:1}). Moreover, to ensure fairness, we also conducted harmonized comparisons where all methods were trained under the same learning rate $1 \times 10^{-4}$) as presented in \cref{app: table 5}, and our method consistently outperformed the baselines. These additional results confirm that our reported superiority is not attributable to hyperparameter selection.

\begin{table}[htb]
  \centering
    \caption{ \raggedright Performance of adopted methods under a learning rate value of $1 \times 10^{-4}$}
\small{
  \begin{tabular}{@{}lcccc@{}}
    \toprule
    Metrics     & CusDiff & DisenDiff & BAS & Ours \\
    \midrule
    \multicolumn{5}{@{}l}{\textit{Single concept}} \\
    CLIP-I$\uparrow$ & 0.528 & 0.551 & \underline{0.564} & \textbf{0.576} \\
    CLIP-T$\uparrow$ & \textbf{0.185} & \underline{0.184} & 0.177 & \underline{0.184} \\
    DINO$\uparrow$   & 0.646    & 0.654   &  \underline{0.662}  &  \textbf{0.703}   \\
    \midrule
    \multicolumn{5}{@{}l}{\textit{Concept group}} \\
    CLIP-I$\uparrow$ & \underline{0.565} & 0.555 & 0.547 & \textbf{0.584} \\
    CLIP-T$\uparrow$ & \textbf{0.211} & 0.201 & 0.195 & \underline{0.209} \\
    DINO$\uparrow$   & 0.648    & \underline{0.667}  &  0.652  &  \textbf{0.723}   \\
    \bottomrule
  \end{tabular}
  }
  \label{app: table 5}
\end{table}

\begin{figure}[htb]
    \centering
    \includegraphics[width=1.0\linewidth]{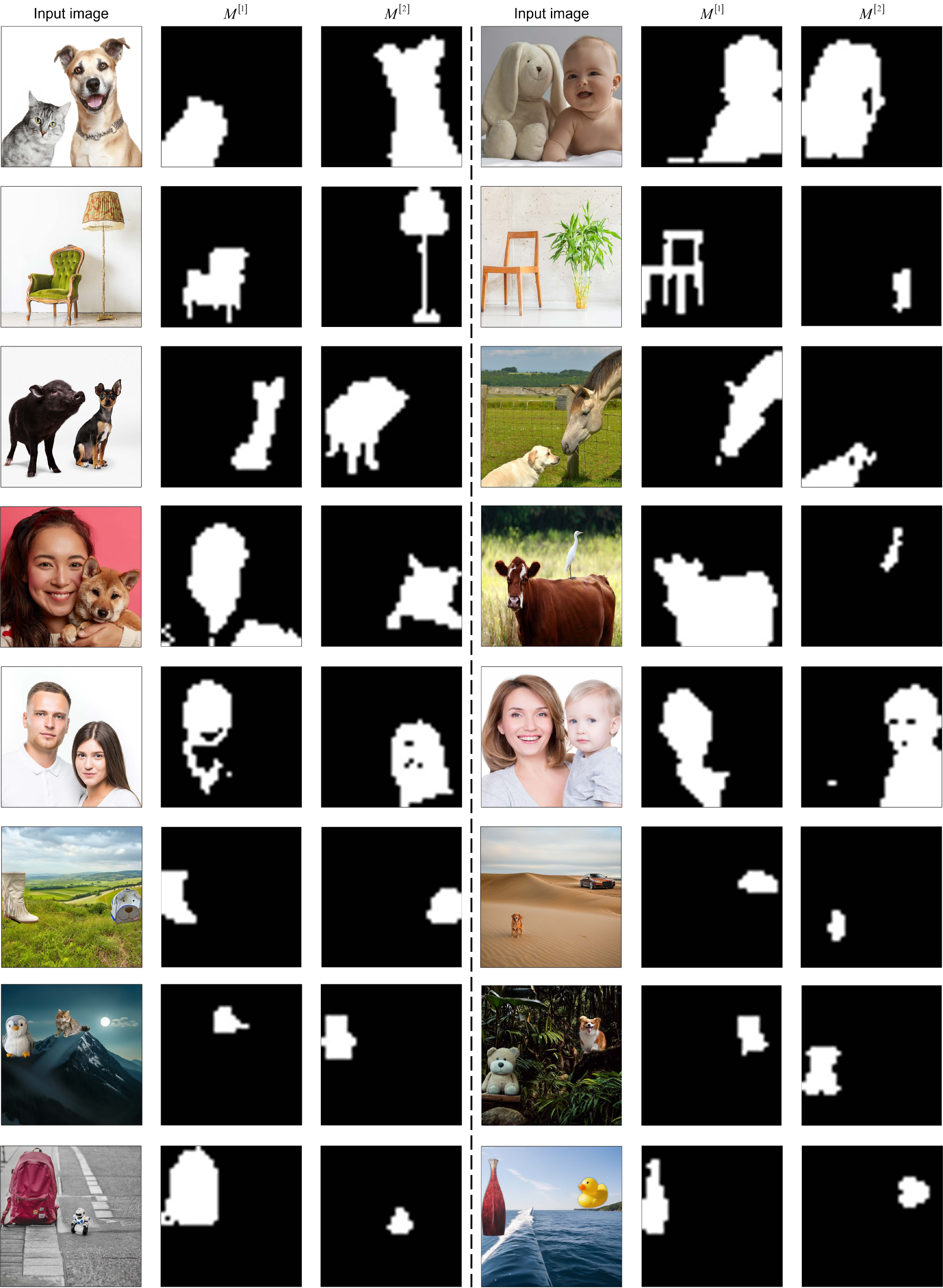}
    \caption{Initial masks for each dataset created by our method}
    \label{app:fig5}
\end{figure}

\subsection*{Mask initialization and sampling ratio determination}
\label{add_results}

In our method, we use a single step to initialize the masks for each concept in the dataset. To present the effectiveness of our method on mask initialization, we present the initial masks for each dataset in \cref{app:fig5}.

Our method tends to present stronger attention on the humans themselves than other parts of humans, such as the long hair and cloth, as illustrated by the woman's mask in the ``woman \& dog" dataset and the mother's mask in the ``mother \& child" dataset. Aside from these, our method successfully creates accurate masks for other datasets.

\end{document}